\documentclass[sigconf]{acmart}

\AtBeginDocument{%
  \providecommand\BibTeX{{%
    \normalfont B\kern-0.5em{\scshape i\kern-0.25em b}\kern-0.8em\TeX}}}

\usepackage{color} 
\usepackage{xspace} 
\usepackage{subfigure}
\usepackage{amsfonts}
\usepackage{booktabs}
\usepackage{multirow}
\usepackage{enumitem}
\usepackage{bm}
\usepackage{amsmath}
\usepackage{natbib}
\usepackage{tikz} 
\usepackage{ulem}
\usepackage{color}
\usepackage{graphicx}

\usepackage[ruled,linesnumbered]{algorithm2e}

\newcommand{\zhou}[1]{{\color{black}#1}} 
\newcommand{\licom}[1]{{\color{black}#1}} 
\newcommand{\liu}[1]{{\color{black}#1}} 
\newcommand{\pending}[1]{{\color{black}#1}} 
\newcommand{\revise}[1]{{\color{black}#1}} 
\newcommand{\hide}[1]{}

\newcommand{\B}[1]{{\bfseries #1}}
\newcommand{\model}{\textsf{SEENet}\xspace}
\newcommand{\gnn}{\textsc{SEConv}\xspace}
\newcommand{\emb}{\textit{ADE}\xspace}
\newcommand{\sconv}{\textit{RS-AGG}\xspace}
\newcommand{\tconv}{\textit{SE-Prop}\xspace}
\newcommand{\ssl}{\textsc{SE-SSL}\xspace}

\newcommand{\problem}{\textsf{Trial}\xspace}

\newcommand{\neighborset}{\mathcal{N}^{2}_{t}(v_i,r_1\rightarrow r_2)}
\newcommand{\graph}{\mathcal{G}}
\newcommand{\nodeset}{\mathcal{V}}
\newcommand{\edgeset}{\mathcal{E}}
\newcommand{\coord}{\mathbf{L}}

\newcommand{\ocat}{\oplus}

\copyrightyear{2023} 
\acmYear{2023} 
\setcopyright{acmlicensed}

\acmConference[KDD '23]{Proceedings of the 29th ACM SIGKDD Conference on Knowledge Discovery and Data Mining}{August 6--10, 2023}{Long Beach, CA, USA}

\acmBooktitle{Proceedings of the 29th ACM SIGKDD Conference on Knowledge Discovery and Data Mining (KDD '23), August 6--10, 2023, Long Beach, CA, USA}

\acmPrice{15.00}
\acmDOI{10.1145/3580305.3599440}
\acmISBN{979-8-4007-0103-0/23/08}

\begin{document}

\title{Multi-Temporal Relationship Inference in Urban Areas}

\author{Shuangli Li}
\authornote{This work was done when the first author was an intern at Baidu Research under the supervision of Jingbo Zhou.}
\affiliation{%
  \institution{School of Computer Science and Technology, University of Science and Technology of China \\  Baidu Research}
  \country{}
}
\email{lsl1997@mail.ustc.edu.cn}

\author{Jingbo Zhou}
\authornote{Corresponding authors.}
\affiliation{%
  \institution{Business Intelligence Lab, \\ Baidu Research}
  \streetaddress{}
  \city{}
  \country{}
}
\email{zhoujingbo@baidu.com}
\orcid{0000-0003-2677-7021}

\author{Ji Liu}
\affiliation{%
  \institution{Big Data Lab,\\ Baidu Research}
  \country{}
}
\email{jiliuwork@gmail.com}

\author{Tong Xu}
\affiliation{%
  \institution{School of Computer Science and Technology, University of Science and Technology of China \& State Key Laboratory of Cognitive Intelligence}
  \country{}
}
\email{tongxu@ustc.edu.cn}

\author{Enhong Chen}
\affiliation{%
  \institution{Anhui Province Key Laboratory of Big Data Analysis and Application, University of Science and Technology of China \& State Key Laboratory of Cognitive Intelligence}
  \country{}
}
\email{cheneh@ustc.edu.cn}

\author{Hui Xiong}
\authornotemark[2]
\affiliation{%
  \institution{The Thrust of Artificial Intelligence, The Hong Kong University of Science and Technology (Guangzhou) \\ The Department of Computer Science and Engineering, The Hong Kong University of Science and Technology}
  \country{}
}
\email{xionghui@ust.hk}


\renewcommand{\shortauthors}{Li and Zhou, et al.}

\begin{abstract} \label{sec-abstract}

\zhou{Finding multiple temporal relationships among locations can benefit a bunch of urban applications, such as dynamic offline advertising and smart public transport planning.}
While \zhou{some} efforts have been made on finding static relationships among \zhou{locations}, little attention is focused on studying \zhou{time-aware} location relationships. Indeed, abundant location-based \zhou{human} activities are time-varying and the availability of these data enables a new paradigm for understanding the dynamic relationships \zhou{in a period} among connective locations. To this end, we propose to study a new problem, namely \textit{multi-Temporal relationship inference among locations} (\problem~\zhou{for short}), where the major challenge is how to integrate dynamic \licom{and geographical} influence under the relationship sparsity constraint. Specifically, we propose \zhou{a} solution \zhou{to} \problem with a graph learning scheme, which includes a spatially evolving graph neural network (\model) with two collaborative components: \licom{spatially evolving graph convolution module (\gnn) and \zhou{spatially evolving} self-supervised learning strategy (\ssl).} \gnn performs the intra-time aggregation and inter-time propagation \licom{to capture the multifaceted spatially evolving contexts from the view of location message passing}\hide{with preserving distant-range spatial and dynamic dependencies simultaneously, as well as involving multi-temporal relational information}. \zhou{In addition}, \ssl designs \pending{ time-aware self-supervised} learning tasks in a global-local manner with \zhou{additional evolving constraint} to enhance the location representation learning and further handle the relationship sparsity. Finally, experiments on four real-world datasets demonstrate the superiority of our method over \zhou{several} state-of-the-art approaches.

\end{abstract}




\maketitle

\section{Introduction} \label{sec-introduction}

\begin{figure}[t]
\centering
\includegraphics[width=1.\columnwidth]{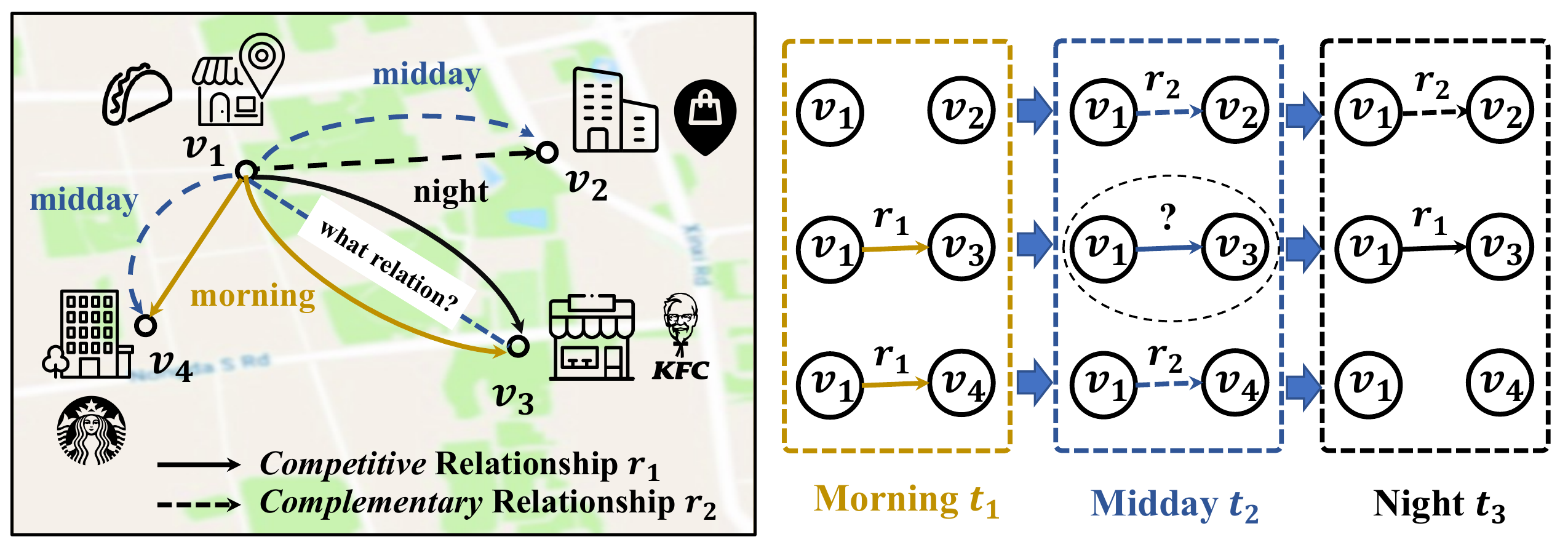}
\vspace{-8mm}
\caption{An illustration of multi-temporal relationships around the \hide{target }fast-food restaurant location $v_1$. \pending{The pair ($v_1$, $v_{2}$) with \textit{complementary} relationships tend to be visited by common users at night, while another pair ($v_1$, $v_3$) with \textit{competitive} relationships provide similar service\hide{ and only one of the pair is chosen to visit} at the specific time.}}
\label{fig-toy-example}
\vspace{-4ex}
\end{figure}
\begin{figure}[t]
\centering
\subfigure{
    \label{analysis-time}
    \includegraphics[width=0.48\columnwidth]{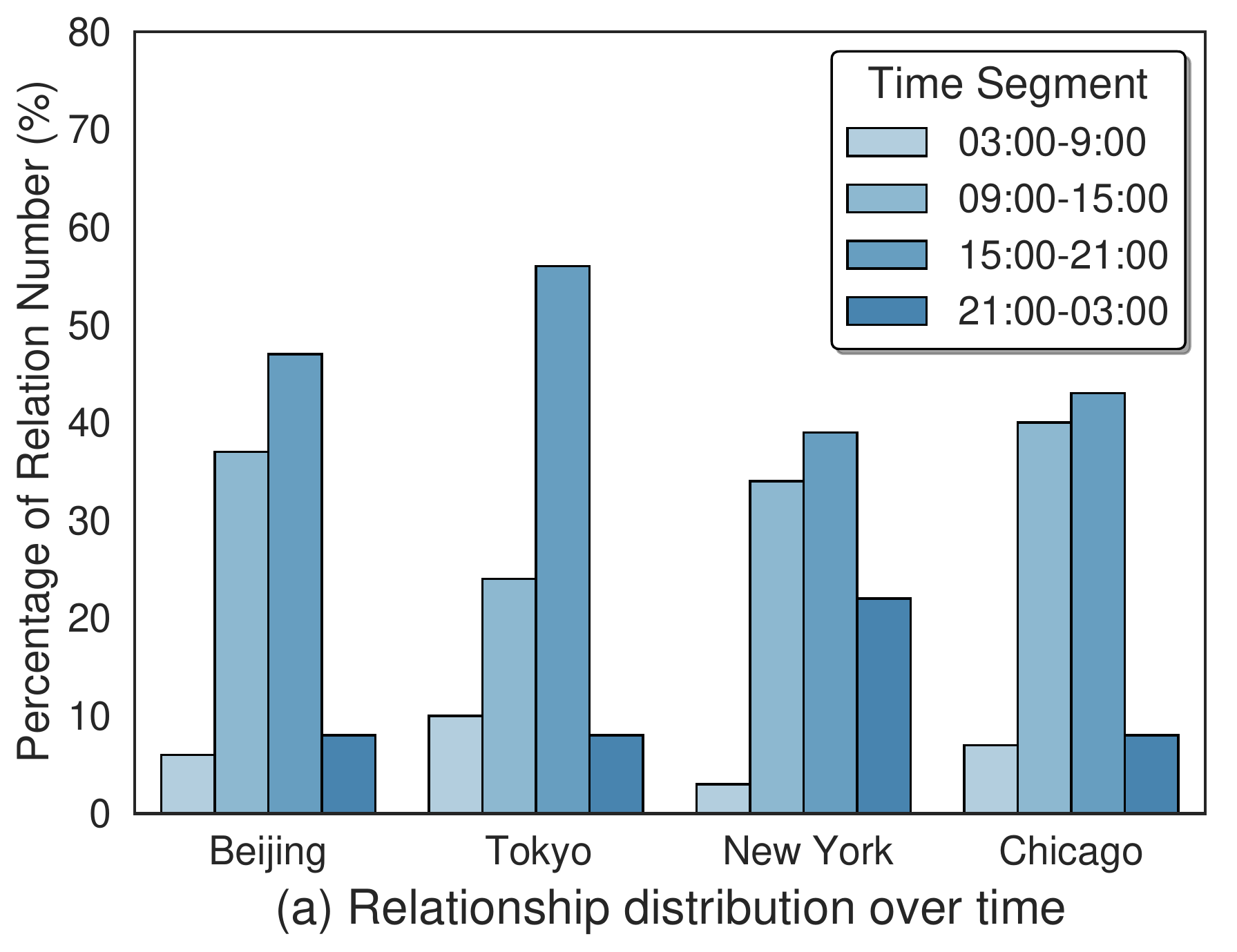}}
  \subfigure{
    \label{analysis-dist-time} 
    \includegraphics[width=0.49\columnwidth]{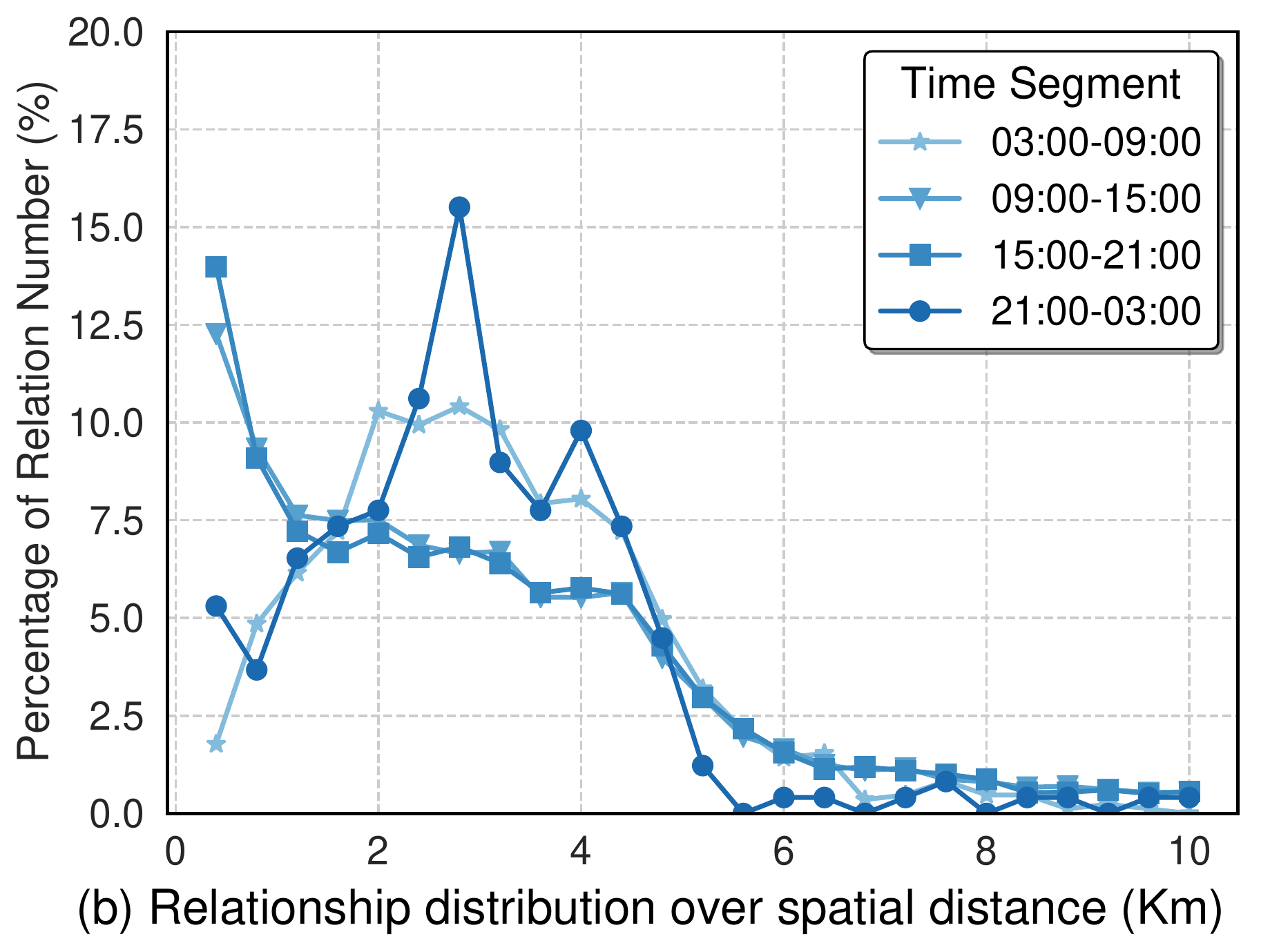}}
\vspace{-8mm}
\caption{Characteristics of \pending{multi-temporal} relationships.}
\label{fig-analysis}
\vspace{-6mm}
\end{figure}

\zhou{
With the widespread availability of human mobility data, discovering location relationships in urban areas has received significant research interest in recent years \cite{liu2022developing, zhou2021competitive, chen2021points, li2020competitive, valkanas2017mining}. Finding the location relationship can help to reveal urban patterns, which definitely benefits intelligent urban management \cite{xiao2022contextual} and finally promotes the urban business economy \cite{liu2022developing}. However, most existing works focus on analyzing static relationships among locations from a spatial perspective, and few known methods study time-aware location relationships from a temporal perspective.
Whereas, the human mobility data is always time-correlated \cite{yuan2013time, li2017time, zhou2018early, jiang2022time}, \textit{e.g., visiting restaurants at noon and visiting bars at night}. Hence, here we present to investigate a new paradigm for understanding such dynamic multiple relationships between locations in a period, which has been largely overlooked in previous studies.

To this end, we propose to study a problem of \textit{multi-\textbf{T}emporal \textbf{r}elationship \textbf{i}nference \textbf{a}mong \textbf{l}ocations} (we call \problem for short), which is of great importance in many urban application scenarios. The goal of \problem is to recover plenty of missing relationships across multiple time segments from the constructed location graph in an urban area. In other words, we aim at mining location connections at different time segments (e.g., morning or evening). Solving \problem can facilitate urban intelligence in many application domains, such as dynamic business advertisements\cite{zhang2020geodemographic}, urban resource planning \cite{gao2023dual,liu2021knowledge}, and knowledge-enhanced location recommendation \cite{wang2021spatio,luo2020spatial}.

Here we further present two examples to demonstrate our motivation for \problem. At first, given a target region location (e.g., a \pending{residential district}), the most relevant regions for this district may be different from morning to evening. After inferring multi-temporal relationships for this district, the urban manager can adaptively plan public transport at different times in a day. Second, let us consider another scenario of urban business as illustrated in Figure \ref{fig-toy-example}. On the one hand, the relationship in an urban area may only exist at specific times due to the evolving daily activities of users \cite{yuan2013time}. In this case, focusing on time-specific correlated locations can help business owners to dynamically optimize their advertisement strategy \cite{mahdian2015algorithmic} at different times. On the other hand, we find the relationship can be diverse in a day. Taking Figure \ref{fig-toy-example} as an example, the cafe $v_1$ (i.e., Starbucks) is competitive with the restaurant $v_2$ (i.e., \revise{Taco Bell}) since both provide breakfast in the morning, while they become complementary at midday due to different services. Therefore, the user experience can be also improved with the discovered multi-temporal location pairs to enhance the time-aware location recommendation \cite{lian2020geography,wang2019knowledge,chen2022building}. Hence, it is critical to effectively understand multi-temporal relationships among locations\hide{ in an urban area}.

Existing methods for relationship mining cannot handle \problem problem effectively since the temporal and geographical factors are rarely considered in a unified manner. Extensive studies have been conducted for relationship mining in other domains, e.g., business analysis \cite{zhang2020large,li2020competitive} or e-commerce \pending{relationship} inference \cite{mcauley2015inferring,liu2020decoupled,liu2021item}. These methods usually cannot be directly used for location relationship mining. Recently, remarkable advancements in graph neural networks (GNNs) \cite{kipf2017semi,vashishth2020composition,wu2020comprehensive} have shown {the} powerful capacity for relationship graph learning and achieved promising results. These GNN-based relationship prediction studies attempt to extend the effective graph message-passing procedure by learning relational correlations \cite{liu2020decoupled,liu2021item} or spatial dependencies \cite{li2020competitive} for location context understanding. The most recent \hide{GNN }method \cite{chen2021points} further adopts attentive aggregation to handle multiple relationships in the urban area. However, all previous works do not fully consider unique attributes of our \problem, which fail to examine two research challenges {in} modeling multi-temporal location relationships as follows.
}

\textbf{\textsc{Challenge 1:}} \textit{How \zhou{to} capture the relational dynamics under \zhou{temporal and} geographical contexts\liu{?}} According to the statistical analysis in Figure \ref{analysis-time}, the distribution of the location relationship varies in a day, indicating the dynamic influence plays a crucial role. \zhou{Moreover,} further investigation in Figure \ref{analysis-dist-time} shows that the relational dynamics is correlated with geography information in the urban area. The geographical context may have different influences on locations at different times. Although some dynamic GNNs \cite{pareja2020evolvegcn,you2022roland} for link prediction are designed for multi-temporal graph structures, they fail to incorporate the relational and spatial evolving patterns. How to communicate the dynamic correlations with geographical factors in an effective way remains a unique research challenge. \pending{Note that spatial-temporal GNN focusing on graph-based time series analysis (e.g., traffic forecasting) is another \zhou{substantially} different research problem \cite{sahili2023spatio,wang2020deep}, which is not applicable to the linking problem of \problem focusing on dynamic relationships.}

\textbf{\textsc{Challenge 2:}} \textit{How \zhou{to} deal with the relationship sparsity with spatial and dynamic influences\liu{?}} Since the precious relationship between locations is derived from limited and conditional user behavior data, the multi-temporal relationship is more scarce \cite{liu2021item}. Therefore, the learning process of GNN suffers from data sparsity with insufficient context information in aggregation. Most existing graph self-supervised learning methods \cite{velickovic2019deep,zhu2021graph,li2022geomgcl} only pay attention to learn\liu{ing} from the augmented graph structures without considering spatial and dynamic influences for relational modeling, thus leading to suboptimal performance. How to take advantage of sparse relationships in the urban area is another notable challenge.
\hide{The final issue is how to tackle the data sparsity problem since the time-oriented relationship is much sparse \cite{liu2021item}.} 

In this paper, we propose a \textbf{S}patially \textbf{E}volving graph n\textbf{E}ural \textbf{Net}work (\model) tailored for inferring  multi-temporal relationships among locations. To address the above challenges, we design the framework from two perspectives: \hide{1) \liu{M}ulti-\liu{S}lot \liu{S}patial \liu{G}raph learning module (\gnn) for message passing-level modeling, and 2) \liu{S}patially \liu{E}volving \liu{S}elf-\liu{S}upervised \liu{L}earning scheme (\ssl) for training-level modeling.}\licom{spatially evolving graph convolution (\gnn) for message passing-level modeling, and 2) spatially evolving self-supervised learning (\ssl) for training-level modeling.} Firstly, aiming at the \textbf{first challenge}, the proposed graph learning procedure \gnn is equipped with intra-time aggregation and inter-time propagation. \pending{The key idea is to identify the spatial and evolving influence from multifaceted locations with considering non-local and cross-time neighbors.} Specifically, the intra-time learning process performs the \licom{second-order aggregation} to preserve \licom{non-local geographical and relational}\hide{distant-range} dependencies at each specific time. On the other hand, the inter-time learning process further \hide{utilizes}propagates the \licom{multi-temporal information to capture the spatially evolving context}\hide{ via the dynamic context-aware interaction layer} across adjacent time segments. Moreover, \ssl is devised to deal with the \textbf{second challenge} of relationship sparsity, which adopts the spatial information maximum strategy from \liu{a} global regional view and the additional evolving constraint from \liu{a} local relational view. By this means, \model can enhance the representation learning for locations with incorporating both spatial distribution and dynamic patterns in a self-supervised manner. The major contributions of this paper are summarized as follows.
\begin{itemize}[leftmargin=*,topsep=3pt]
    \item To the best of our knowledge, this is the first work to \pending{investigate} the problem of \textit{multi-temporal \hide{location} relationship inference} \zhou{among locations} for various valuable scenarios, which studies time-specific relationships in urban areas at a fine-grained level.
    \item We propose a novel spatially evolving graph neural network named \model with collaborative designs for relationship learning among locations, which can capture the geographical and dynamic influence through an intra-time and inter-time spatially evolving graph convolution as well as an effective evolutionary self-supervised learning task.
    \item We conduct extensive experiments on four real-world datasets\hide{ for multi-time relationship discovery}, which demonstrates the superiority of \model.\hide{ towards representative baselines.}
\end{itemize}
\vspace{-2.5ex}

\section{Related Work} \label{sec-related}
In this section, we review the previous literature from two perspectives: topic-related \textit{relationship mining and inference}, and closely technology-related \textit{graph neural networks}.
\vspace{0.7ex}

\B{Relationship Mining and Inference.} 
Mining valuable relationships \hide{is a long-standing problem}has attracted increasing attention from both academia and industry \cite{mcauley2015inferring,werle2022competitor}. In the early stage, most of the previous works aim at inferring the precious relationship from content information \cite{valkanas2017mining} (e.g., textual reviews and descriptions on the web), while expert knowledge is required to design linguistic rules \cite{li2006cominer,jindal2006identifying} or graphical analysis models \cite{yang2012mining}. Another line of work seeks to apply deep learning-based techniques to analyze various relationships. Some explore applying the linked auto-encoder \cite{rakesh2019linked} and graphlet mining \cite{wang2018path,zhou2021competitive} for product-oriented or competitor-oriented applications. Considering the natural inadequacy in learning relational graph structures of these domain-specific methods, some recent works further propose to develop powerful graph learning approaches for relationship discovery. From the perspective of relational dependency learning, DecGCN \cite{liu2020decoupled} designs the graph structural integration mechanism for decoupled representation learning, which can detect the mutual influence between different relationships. The recent IRGNN \cite{liu2021item} further incorporates the multi-hop complex relationships to alleviate the sparsity issue. However, spatial dependencies between nodes are omitted which encourages researchers to propose geography-based graph methods. Therefore, from the perspective of spatial context modeling in urban area, the fine-grained distance distribution \hide{around the POI node }is captured in DeepR \cite{li2020competitive} with spatial adaptive graph convolutions. More recently, PRIM \cite{chen2021points} combines self-attentive spatial context extractor for multiple relation types. However, the important relational dynamics is neglected all along. In this paper, we focus on the challenging multi-temporal relationship inference among locations to fill the research gap.

\vspace{0.5ex}
\B{Graph Neural Networks.} Recent years have witnessed the rapid growth of graph neural networks (GNNs), which exhibit a strong ability in learning structural relationships \cite{liu2020decoupled,li2020competitive,Xu2023Multimodal,Xiao2021Watcher,hao2021demand,li2021structure}. \pending{According to the unique challenging properties of location relationships as introduced before, the technically corresponding GNN models fall into three mainstreams.} Firstly, a number of GNNs perform diverse message-passing schemes to capture rich context information from graph structures, such as considering edge types for multiple relational semantics \cite{vashishth2020composition} and spatial attributes \cite{chen2021points,li2020competitive}. Moreover, \hide{since only aggregating the neighboring signals is inadequate for the complex graph, Many}some efforts have been devoted to studying expressive high-order GNN methods, including mixing neighboring features at various distances in Mixhop \cite{abu2019mixhop}, distinguishing non-local topological structures with the random walk \cite{eliasof2022pathgcn} or attention-guided sorting \cite{liu2021non}. Secondly, graph self-supervised learning methods focus on designing augmented strategies to tackle data scarcity, which utilizes contrastive learning \cite{zhu2021graph} or meaningful tasks \cite{velickovic2019deep} on graphs. The most recent RGRL \cite{lee2022relational} also leverages the relationship information with preserving global and local similarity. Nevertheless, these methods tend to lose effectiveness without considering the spatial and evolving characteristics of relationships\hide{the self-supervised learning process}. Finally, \hide{despite the success of spatial-temporal GNNs,} integrating relational dynamics is also important \pending{for snapshot-based graph learning}. Along this line, EvolveGCN \cite{pareja2020evolvegcn} is proposed to recurrently updates the GNN weights for dynamic link prediction. Although the recent ROLAND \cite{you2022roland} takes a further step to involve hierarchical states over time, one-sided dynamic information is still not sufficient. It is noteworthy that the irrelevant spatial-temporal GNNs can not be applied in location relationship learning for comparison. Because most of them are basically designed for time series forecasting with specific sequential values (e.g., traffic flow or weather conditions) \cite{wang2020deep}, which is intrinsically different from our target of multi-time link prediction. Therefore, we aim to develop an adaptive graph neural network to preserve both spatial and dynamic dependencies simultaneously for location relationship inference.
\vspace{-0ex}
\section{Preliminaries}\label{sec-pre}
In this section, we first present the concept of the dynamic location graph, and then formalize the problem of multi-temporal relationship inference among locations (\problem for short).

\vspace{-0.5ex}
\begin{definition}
\B{(Time-specific Location Relationship)}. As stated above, relationships between locations are dynamic in a day. Thus, the complex location relationships should be naturally decomposed into multi-temporal segments to meet the unique daily dynamic characteristic of locations.\hide{To deal with such unique daily dynamic characteristic of POIs, we decompose complex POI relationships with multi-time segments, which are pre-defined by domain experts.} Formally, the \hide{well-designed}\licom{pre-defined} set of time segments is denoted as $\mathbb{T}=\{t_1,t_2,...,t_T\}$, where all $T$ time segments form a whole day. Accordingly, the location relationship set $\mathcal{R}$ contains multiple time-specific relationships among locations. In the following sections, the mentioned \textit{relationship} also stands for \textit{location relationship}. We use $t\in\mathbb{T}$ to generally represent a certain time, while $r\in\mathcal{R}$ refers to a given relationship. \hide{is decomposed into multiple time-specific relationships $\mathcal{R}=\bigcup\limits_{t=1}^N R^{(T_t)}$.}
\end{definition}
\vspace{-0.ex}
\begin{definition}
\B{(Dynamic Location Graph).} In this work, the multi-temporal relationships are organized as a fine-grained dynamic location graph $\mathcal{G}=\{G^{(t)}|G^{(t)}=(\nodeset, \edgeset^{(t)}, \coord), t\in \mathbb{T}\}$, where $\nodeset=\{v_1,v_2, ..., v_N\}$ is the set of location nodes with the spatial coordinates $\coord \in \mathbb{R}^{2\times N}$. The dynamic graph $\mathcal{G}$ is composed of $T$ time-specific graphs and contains different relationship edges at different time segments, which is denoted as $\edgeset=\edgeset^{(t_1)}\cup...\cup\edgeset^{(t_T)}$. The relational edge $e_{i,j,r}^{(t)}=(v_i,v_j,r,t)$ represents there exists the relationship $r\in\mathcal{R}$ between $v_i$ and $v_j$ at time \hide{segment }$t$.
\end{definition}
\vspace{-0.5ex}
Since the relational location graph $\mathcal{G}$ is usually sparse and most valuable relationships are absent, our target is to learn from $\mathcal{G}$ and \liu{discover} all meaningful relationships at different time segments. We formally define the problem as below:
\vspace{-0.ex}
\begin{figure*}[t]
\centering
\includegraphics[width=.90\textwidth]{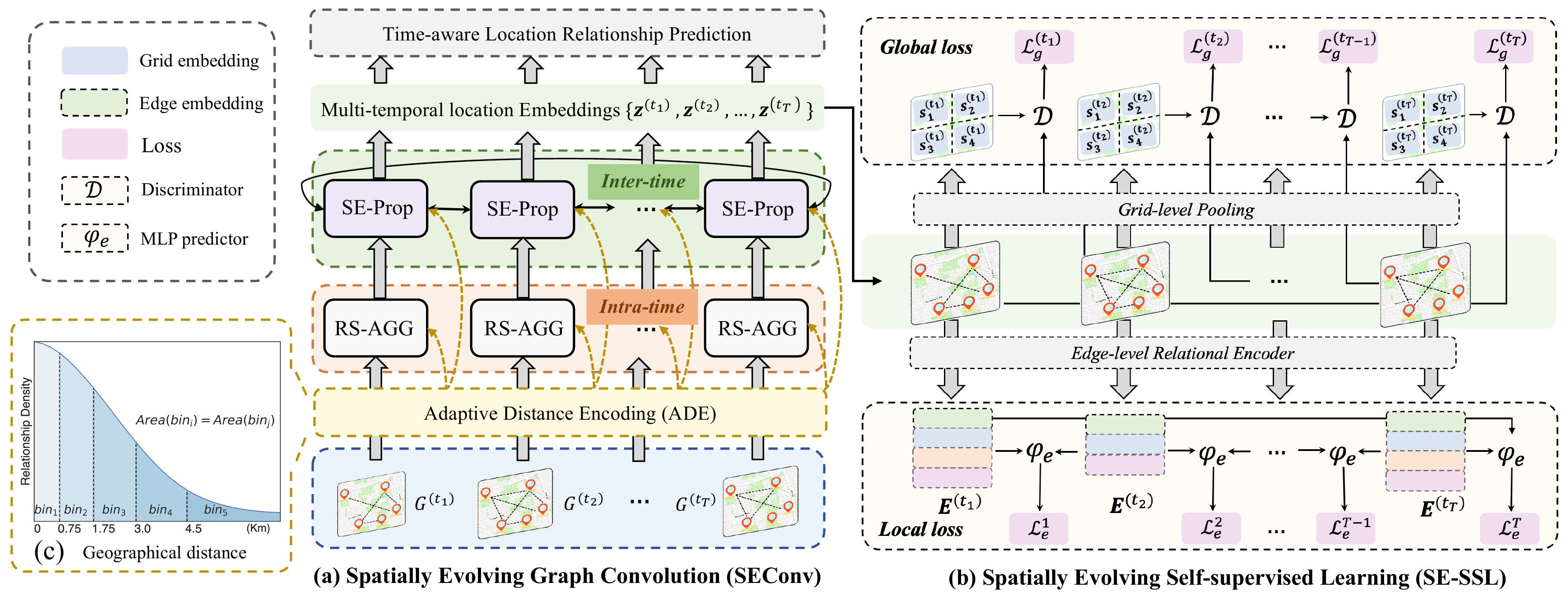}
\vspace{-4mm}
\caption{Illustration of the proposed \model framework for dynamic location graphs. 
}
\label{fig-model}
\vspace{-5mm}
\end{figure*}
\begin{definition}
\B{(Multi-Temporal Relationship Inference).} \hide{Given}\licom{For} the dynamic location graph $\mathcal{G}=\{G^{(t_1)},\cdots,G^{(t_T)}\}$ associated with \hide{multi-time segments}\licom{spatial locations $\coord$,} \zhou{the} problem of \problem aims to jointly learn a model $\mathcal{F}(\{G^{(t_1)},\cdots,G^{(t_T)}\})$ to map the location node set $\nodeset$ into multiple embeddings $\{\mathbf{z}^{(t_1)},\cdots,\mathbf{z}^{(t_T)}\}$ at multi-temporal segments. \hide{such that the possibility score for each relationship $r$ between a location pair $(v_i,v_j)$ can be estimated by prediction function $p^{(t)}_r(\mathbf{z}^{(t)}_i,\mathbf{z}^{(t)}_j)$ at each time segment $t$.}\licom{Then, given a location pair $(v_i,v_j)\in\mathcal{V}\times\mathcal{V}$ at the time $t\in\mathbb{T}$, the possibility score for each relationship $r$ can be estimated by the prediction function $p^{(t)}_r(\mathbf{z}^{(t)}_i,\mathbf{z}^{(t)}_j)$.} Therefore, we can \liu{discover} all potential location relationships at multiple times.
\end{definition}
\vspace{-1.6ex}

\section{Model Framework}\label{sec-model}
In this section, we present the \underline{S}patially \underline{E}volving graph n\underline{E}ural \underline{Net}work (\model) model, which learns from \zhou{multi-temporal dynamic and geographical} correlations in an end-to-end manner. As illustrated in Figure \ref{fig-model}, the overall framework first \hide{handle}take\liu{s} the dynamic location graph $\graph$ in the temporal format as input. Our proposed \model is equipped with a \licom{spatially evolving graph convolution module} (\gnn) to incorporate the evolving context along with the comprehensive spatial influence. After obtaining the representations of locations by \gnn, the spatially evolving self-supervised learning module (\ssl) is devised to enhance the model's capability of learning dynamic relational patterns through well-designed training tasks on sparse location graphs.\hide{across time segments} Finally, we utilize the pre-trained model after \ssl for our problem of \problem\hide{ stage at each time segment}.
\vspace{-1ex}

\subsection{Spatially Evolving Graph Convolution}
\zhou{The intrinsic evolving correlations associated with spatial contexts are critical to time-specific location representation learning.} In past years, \hide{graph neural networks (GNNs)} GNNs \cite{kipf2017semi} have shown the superiority on processing relational graph structures for Points-of-Interest \cite{li2020competitive,chen2021points} or items \cite{liu2020decoupled,liu2021item}. These GNN methods mainly focus on topological structures with learning spatial dependencies or relational semantics on the single static graph, which fails to deal with multi-temporal relations. \hide{The intrinsic evolving correlations associated with spatial contexts are critical to time-specific location representation learning.}

\hide{Considering the unique attributes of the dynamic location graph,} \zhou{To this end,} we develop the \textbf{S}patially \textbf{E}volving graph \textbf{Conv}olution (\gnn) to capture both spatial location context and dynamic correlations across \hide{multiple time segments}time. The key idea of \gnn derives from two perspectives for complex context modeling: \textit{intra-time} non-local relational interactions and \textit{inter-time} spatially evolving interactions. 

Since the geography information plays a crucial role in location relationship\hide{ as presented in Figure \ref{fig-analysis}}, \zhou{before introducing the main components of \gnn,} we first manage to project the scalar distances into informative spatial representations. In view of the varying spatial distribution of different datasets or cities, we \revise{adopt} the \liu{A}daptive \liu{D}istance \liu{E}ncoder (\emb) with distribution-aware embedding mechanism \cite{xu2022g2net}. We first calculate all distances based on location's coordinate matrix $\coord$ in the relational graph $\graph$, then the statistical distance distribution $P(x)$ is obtained in Figure \ref{fig-model}(c). Since the scalar distance value only has limited one-dimension information without learning ability, we build the embedding layer to extract the discrete representations. \licom{As illustrated in Figure \ref{fig-model}(c)}, the $P(x)$ distribution is uniformly decomposed into $N_b$ consistent distance-space bins with the constraint of the equal area size under each bin's curve. The boundary list $\bm{B}(N_b)$ for distance bins is calculated as:
\begin{equation}
    \label{eq-embed-1}
    \bm{B}(N_b) = [b_1, b_2, ..., b_{N_b}] \quad \mathrm{s.t.} \ \int_{0}^{b_k}\textit{P}(x)\, dx = \frac{k}{N_b},
\end{equation}

Given a pair of locations $(v_i, v_j)$, we further map the distance $||\coord_i - \coord_j||$ to the bin index $k$ based on the uniform distance boundaries. The discrete representation $\mathbf{d}_{i,j}$ with the bin index $k$ is obtained through the embedding layer:
\begin{equation}
    \label{eq-embed-2}
    \mathbf{d}_{i,j} = \mathrm{Embedding}(k) \quad \mathrm{s.t.}\ \ b_k \leq ||\coord_i - \coord_j|| < b_{k+1},
\end{equation}
where $\coord_i$ is the location coordinate for \hide{POI}$v_i$. The generated distribution-aware distance representations can adaptively imply the overall spatial context from the view of statistical analysis. \hide{ without expert knowledge. }

After that, the spatial distance embedding is integrated into the \textit{\liu{R}elational \liu{S}patial \liu{AGG}regation} (\sconv)  for time-specific non-local dependencies modeling and the \textit{\liu{S}patially \liu{E}volving \liu{c}ontextual} \liu{Prop}agation (\tconv) for multifaceted context modeling, \zhou{\hide{\sconv and \tconv}which will be introduced later.} As shown in Figure \ref{fig-model}(a), the overall intra- and inter-time convolutional process at \hide{the }time \hide{segment} $t$ is defined as:
\begin{equation}
\begin{split}
    \label{eq-layers}
    \mathbf{h}_{i,intra}^{(t)} &= \mathrm{AGG}_t \Big ( \big\{(\mathbf{h}_j, \mathbf{h}_k, \mathbf{d}_{i,k}) \ \big | \ \forall v_j \in \mathcal{N}^{(t)}_i, v_k \in \mathcal{N}^{(t)}_j \big\} \Big ),
    \\
    \mathbf{h}_{i,inter}^{(t)} &= \mathrm{Prop}_t \Big ( \big\{(\mathbf{h}_{j,intra}^{(\tau)},  \mathbf{d}_{i,j}) \ \big | \ \forall v_j \in \mathcal{N}^{(\tau)}_{i}, \tau \in T(t)\big\} \Big ),
\end{split}
\end{equation}
where $\mathbf{h}_i$ is the input location embedding, $\mathbf{h}_{i,intra}^{(t)}$ and $\mathbf{h}_{i,inter}^{(t)}$ are the intra-\hide{time} and inter-time embeddings for location\hide{ node} $v_i$, $\mathrm{AGG}_t$ and $\mathrm{Prop}_t$ are the aggregation and propagation functions at the time $t$ for \sconv and \tconv, respectively. $\mathcal{N}_{i}^{(t)}$ is the neighboring set of $v_i$ at time \hide{segment }$t$. \hide{The function $g^{(t)}$ for \tconv performs inter-time graph convolutions across adjacent multi-time segments $T(t)$ around time $t$.}\licom{$T(t)$ represents multiple adjacent time segments (from $T_{t_1}$ to $T_{t_2}$) around $t$ for the inter-time \hide{graph learning}propagation process.}
\begin{figure}[t]
\centering
\subfigure{
    \label{2hop-dist}
    \includegraphics[width=0.45\columnwidth]{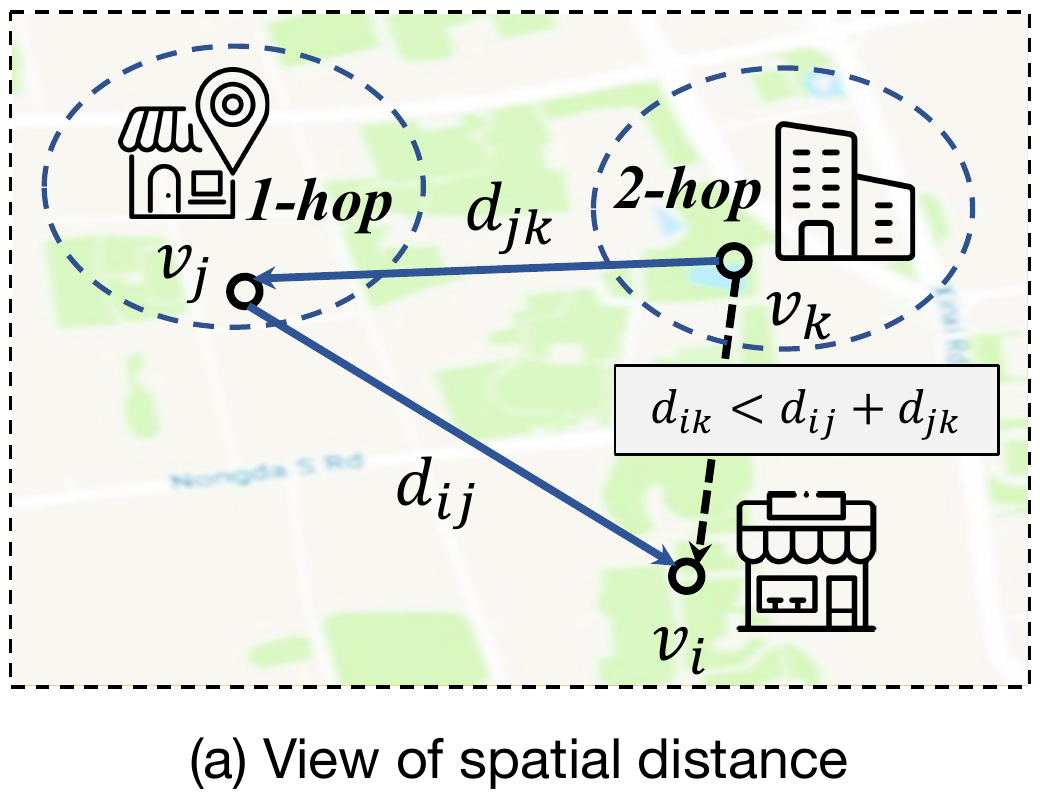}}
  \hspace{-2mm}
  \subfigure{
    \label{2hop-rel} 
    \includegraphics[width=0.435\columnwidth]{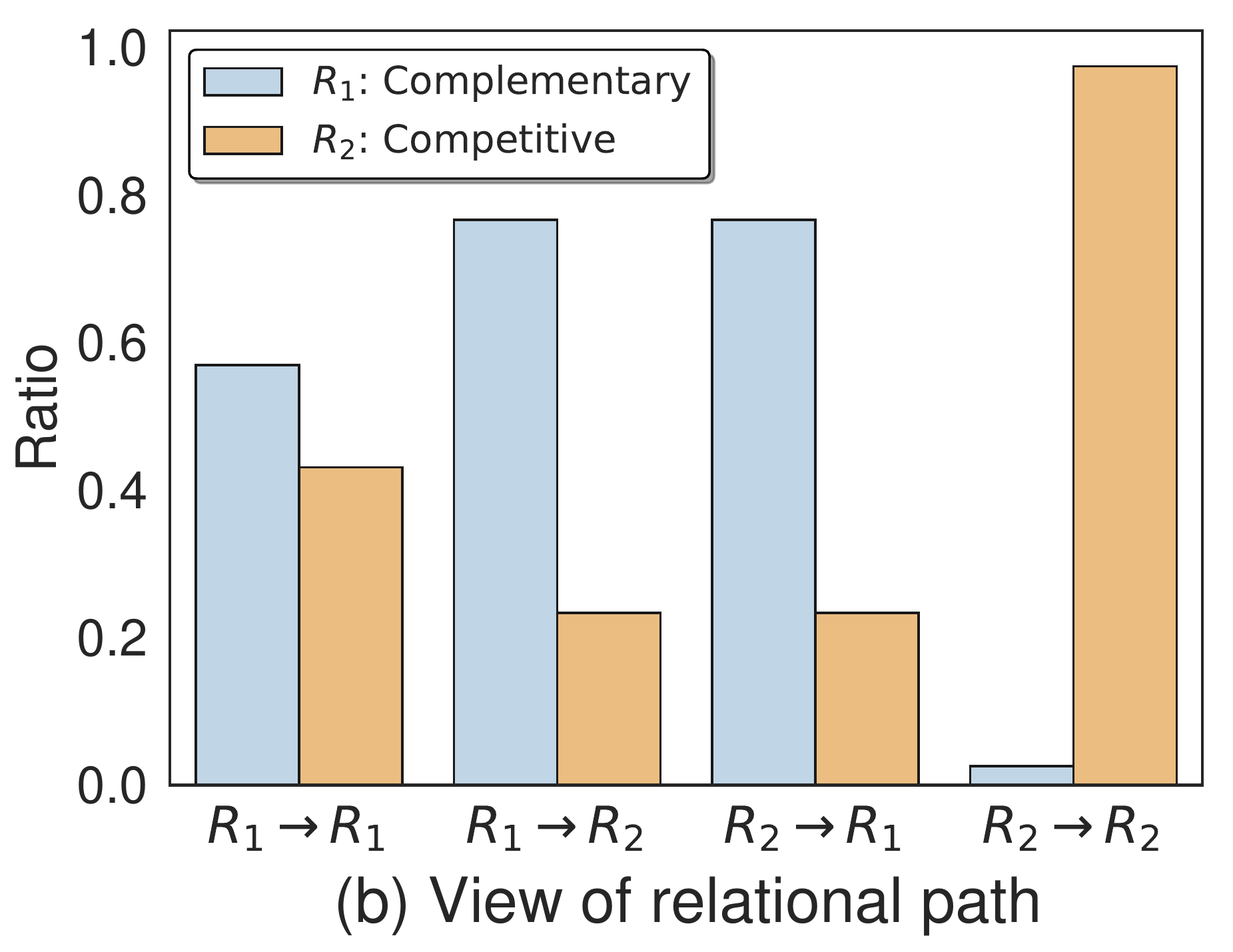}}
\vspace{-5mm}
\caption{Two views of \hide{understanding }second-order dependencies in \sconv.\hide{idea: In fact, the connectivity of POIs more than one hop away also contains rich semantics and geographical dependency, which could be utilized for improved relationship prediction.}}
\label{fig-2hop}
\vspace{-6mm}
\end{figure}

\subsubsection{\B{Intra-time Interaction: Relational Spatial Aggregation (\sconv)}}

\licom{
In the scenario of location relationships, the abundant second-order \hide{contexts}dependency can potentially imply the undiscovered relationships. To meet the needs of such unique location graph learning, the proposed \sconv adopts the second-order message passing architecture with considering the non-local \hide{location }environment in the first stage of intra-time interaction modeling. Different from previous GNN-based relationship learning methods aggregating information from 1-hop local neighbors, the basic intuition of \sconv derives from both spatial and relational views.
}

\B{From the spatial view.} As shown in Figure \ref{2hop-dist}, although the 2-hop neighboring locations are not directly connected with the target location $v_i$, they may be geographically close to $v_i$. As we have investigated in Figure \ref{analysis-dist-time}, the stronger geographical connection (i.e., closer distance) indicates the higher possibility of relationship existence. However, the ordinary 1-hop message passing process can only deliver the spatial distance between neighboring locations. If we simply stack two GNN layers with propagating distance information $(d_{ij},d_{jk})$, the model can not capture the real 2-hop distance $d_{ik}$ in the geographical space (usually $d_{ij}+d_{jk} \ne d_{ik}$ unless three nodes $(v_i,v_j,v_l)$ are spatially collinear). 

\B{From the relational view}. The 2-hop relational path can reflect the meaningful relationship patterns, which is denoted as $v_i \stackrel{R_{i,j}}{\longrightarrow} v_j \stackrel{R_{j,k}}{\longrightarrow} v_k$. As shown in Figure \ref{2hop-rel}, different types of relational combinations can imply different potential relationships, where the relational dependence $R_{i,j}\rightarrow R_{j,k}$ is helpful to reveal the relationship $R_{i,k}$. Distinguishing such path semantics can greatly improve the model's ability to discover more hidden relationships. 

To \licom{utilize}\hide{make use of} the above two characteristics among locations, we incorporate the spatial influence and relational dependence into a unified aggregation network \sconv. \hide{For instance,}\hide{which represents the complex relational dependencies for discovering hidden relationships.}We first define the second-order neighbors with the mixed relational division strategy:
\begin{equation}
\label{eq-2hop-neighbor}
    \mathcal{N}^{2}_{t}(v_i,r_1\rightarrow r_2) = \big\{(v_j,v_k) \ \big| \ \underbrace{v_i \stackrel{R_{i,j}}{\rightarrow} v_j \stackrel{R_{j,k}}{\rightarrow} v_k \models r_1\rightarrow r_2}_{{}{\rm Relational\ path\ constraint}} \big\},
\end{equation}

where $r_1$ and $r_2$ denote a certain relation pair, $\models$ means that the relational path determined by the triple nodes $(v_i,v_j,v_k)$ satisfy the 2-hop relational pattern $r_1\rightarrow r_2$ (i.e., $R_{i,j}=r_1$ and $R_{j,k}=r_2$). The neighboring set $\neighborset$ defines the associated location pair $(v_j,v_k)$ within the second-order range, where the 2-hop neighbor $v_k$ is reachable via the middle 1-hop neighbor $v_j$ \pending{on the graph at time $t$}\hide{at the $t$-th time segment graph}. \licom{Only the pairs connected by at least two different relational paths are included for efficiency.} In this way, we can provide the most crucial evidence for inferring \hide{POI }relationships. 

Then\liu{,} we further propose the \hide{relational interactive}geography-aware relational graph convolutions to preserve multiple dependencies from the complex intra-time connections. This non-local learning scheme is devised to handle the second-order correlations\hide{ among nodes}\hide{ objectives} for each pattern, which can simultaneously aggregate the 1-hop and 2-hop interaction features with the spatial gating mechanism. In general, the \licom{path-specific}\hide{pattern-specific} location representation for $v_i$ is generated as follows:
\licom{
\begin{equation}
\label{eq-relation-1}
\small
    \mathbf{h}_{i,r_1\rightarrow r_2}^{(t)} = \sum_{(v_j,v_k) \in (r_1 \rightarrow r_2)} \Big(\mathbf{W}_{r_1\rightarrow r_2}^{(t)}\mathbf{h}_j^{(t)}+ \Phi_{t}(i,j,k)\cdot\mathbf{W}_{r_1\rightarrow r_2}^{(t)}\mathbf{h}_k^{(t)}\Big),
\end{equation}
}

where $(r_1 \rightarrow r_2)$ is simply short for the constructed second-order neighboring set $\neighborset$, $\mathbf{h}^{(t)}_j$ and $\mathbf{h}^{(t)}_k$ are the input embeddings of 1-hop neighbor $v_j$ and 2-hop neighbor $v_k$ respectively, $\mathbf{W}_{r_1\rightarrow r_2}^{(t)}$ is the weight matrix for the specific relational path pattern $r_1 \rightarrow r_2$ at \hide{the $t$-th time segment}\pending{time $t$}. The proposed operator $\Phi_t$ represents the spatial gating function to determine the distinctive influence of the second-order information, which is formulated as:
\begin{equation}
\label{eq-gate}
    \Phi_t(i,j,k) = \mathrm{sigmoid}\big(\mathbf{a}^T_{t,r_1,r_2}\cdot(\mathbf{s}^{spa}_{t,r_1,r_2}+\mathbf{s}^{rel}_{t,r_1,r_2})\big),
\end{equation}
where $\mathbf{a}^T_{t,r_1,r_2}$ is the trainable parameter for importance weight calculation. Here we take the 1-hop relation semantics and 2-hop geographical impact into account. Since the calculated gating score is devised to reflect the relative significance between 1-hop and 2-hop information, it combines the pairwise relational-based vector $ \mathbf{s}^{rel}_{t,r_1,r_2}$ and spatial-based vector $ \mathbf{s}^{spa}_{t,r_1,r_2}$ from two domain spaces\hide{aspects}:
\begin{equation}
\label{eq-score-1}
    \mathbf{s}^{spa}_{t,r_1,r_2} = \mathbf{W}^{spa}_{t,r_1,r_2}\big[\mathbf{G}_{t,r_1,r_2}\mathbf{d}_{i,j} \ocat \mathbf{W}_{r_1\rightarrow r_2}^{(t)}\mathbf{h}^{(t)}_{k}\big],
\end{equation}
\begin{equation}
\label{eq-score-2}
    \mathbf{s}^{rel}_{t,r_1,r_2} = \mathbf{W}^{rel}_{t,r_1,r_2}\big[\mathbf{W}_{r_1\rightarrow r_2}^{(t)} \mathbf{h}^{(t)}_{j} \ocat \mathbf{W}_{r_1\rightarrow r_2}^{(t)} \mathbf{h}^{(t)}_{k}\big],
\end{equation}
where $\mathbf{W}^{spa}_{t,r_1,r_2}$, $\mathbf{W}^{rel}_{t,r_1,r_2}$, and $\mathbf{G}_{t,r_1,r_2}$ denote learnable weighted matrices, $\ocat$ represent the concatenation operation. $\mathbf{G}_{t,r_1,r_2}$ transforms the spatial distance representation $\mathbf{d}_{i,j}$ \hide{obtained from the adaptive distance encoder }in the \hide{pattern-specific}\licom{relational path-specific} latent space. Under the guidance of dual-factor gated mechanism, the informative second-order aggregation process involves both relational and spatial signals. 

After the relational spatial graph convolution scheme is performed for all hybrid paired patterns $(r_1,r_2)$, we combine all \hide{pattern-specific}\licom{path-specific} location representations obtained from Eq. (\ref{eq-relation-1}) with mean pooling to strengthen the intra-time dependency learning.
\begin{equation}
\label{eq-relation-2}
    \mathbf{h}_{i,intra}^{(t)} = \sum_{(r_1,r_2)\in\mathcal{R}\times\mathcal{R}} \frac{1}{|\mathcal{R}\times\mathcal{R}|} \cdot \mathbf{h}_{i,r_1\rightarrow r_2}^{(t)},
\end{equation}
where $\mathcal{R}$ is the relationship set, $|\mathcal{R}\times\mathcal{R}|$ \pending{is the number of paths}\hide{represents the number of paired relational patterns}.

\subsubsection{\B{Inter-time Interaction: Spatially Evolving Contextual Propagation (\tconv)}} 

\licom{The diversified correlations between locations are also heavily dependent on inter-time interactions. We further propose the spatially evolving propagation layer to capture complex contextual messages, which complements the intra-time spatial aggregation from a dynamic perspective. After updating the location embedding via the inter-time fusion layer, the well-designed \tconv explores to integrate the spatially evolving context among multi-temporal location neighbors for better location relationship learning, since the spatial dynamics at different times can play a great role in relationship mining as introduced before.

Specifically, for each location in dynamic graphs, the semantics of embedding at different segments are distinct\hide{ with the latent evolution}. Location embeddings at adjacent time segments $T(t)$ can provide the sequential latent information, which has the potential effect on the current location relationship at $t$ due to the time continuity. Thus, the inter-time fusion layer over multiple times is first adopted to update the temporal-enhanced location representation \hide{$\widetilde{\mathbf{h}}_{i}^{(t)}$ at the time $t$}:
\begin{equation}
\label{eq-dy-agg}
    \widetilde{\mathbf{h}}_{i}^{(t)} = \sum_{\tau \in [T_{t_1},T_{t_2}]} \frac{1}{T_{t_2}-T_{t_1}} \cdot \mathbf{W}_t\mathbf{h}_{i,intra}^{(\tau)}, \ \ \ T_{t_1}\le t \le T_{t_2},
\end{equation}
where $\mathbf{W}_t$ is the time-specific transformation matrix, the prior time $T_{t_1}$ and the later time $T_{t_2}$ define the duration of aggregated time segments $T(t)$ (we set $T_{t_1}$ and  $T_{t_2}$ as $t-1$ and $t+1$ in practice).

\hide{complicated}Then we further simultaneously leverage the spatial and evolving characteristics among locations in the propagation. As illustrated in Figure \ref{fig-dynamic-example}, different from the classic message passing scheme focusing on pairwise local interactions, the \tconv additionally considers the spatially evolving context when performing the propagation from \hide{a certain}each neighboring location $v_j$ to the target $v_i$. The relationship structure is evolving over time, and a location in the time-specific graph may contain limited connections. Supplying cross-time neighboring nodes can \hide{greatly }enrich the relational environment in the propagation of $v_j \rightarrow v_i$. Therefore, we first utilize the temporal \hide{random }sampling strategy to gather abundant neighbors, which provides the critical evolving information for the message $v_j \rightarrow v_i$: 
\begin{equation}
\label{eq-temporal-set}
    \mathcal{N}_{K}(v_i,v_j) = \{v_{i_1}, ..., v_{i_K} \ | \ v_k = \mathrm{Sampling}(\cup_{T_{t_1}}^{T_{t_2}} \mathcal{N}^{(t)}_i \backslash \{v_j\}) \},
\end{equation}
where $\mathrm{Sampling}$ stands for the random sampling process across time segments. The set $\mathcal{N}_K$ collects $K$-size nearby neighbors of $v_i$ with the repeated sampling from multi-temporal neighboring views, which builds the bridge between temporal-correlated locations from time $T_{t_1}$ to $T_{t_2}$ for \hide{distribution }evolving context construction (e.g., picking up $\{v_{i_1},v_{i_2},v_{i_3}\}$ from $t_1$ to $t_3$ in Figure \ref{fig-dynamic-example}). 

\begin{figure}[t]
\centering
\includegraphics[width=1.\columnwidth]{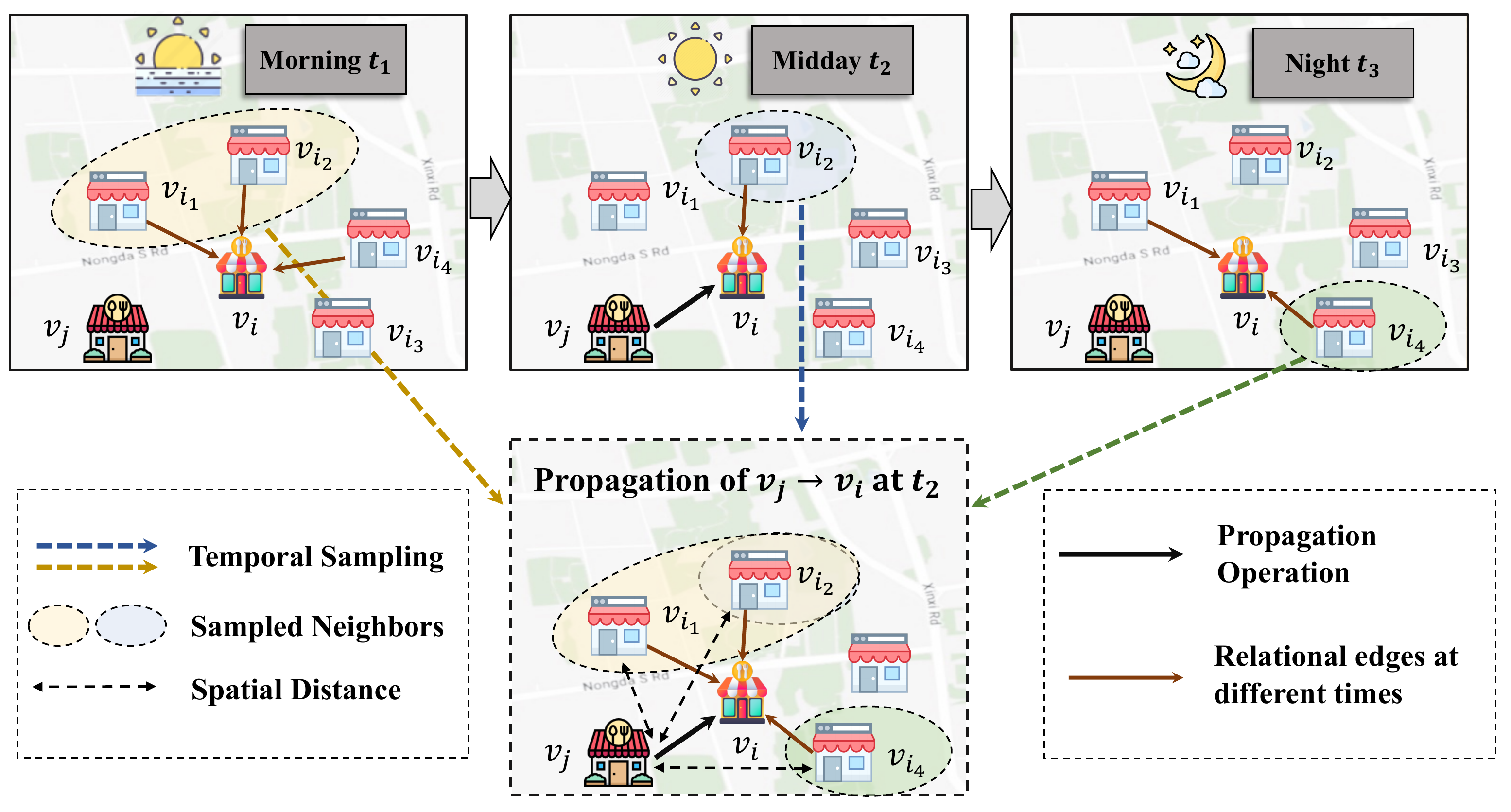}
\vspace{-7mm}
\caption{An illustrated example of spatially evolving context construction with random temporal sampling in \tconv.\hide{ at consecutive time segments. }}
\label{fig-dynamic-example}
\vspace{-6mm}
\end{figure}

Moreover, the sampled locations are spatially distributed around the target interactive pair $v_j \rightarrow v_i$. The geographical knowledge under the evolving influence can offer multifaceted contexts from a comprehensive view. We introduce an informative vector $\overline{\mathbf{C}}_{i,j}^{(t)}$ to extract such spatially evolving context among the edge $v_j \rightarrow v_i$ with considering the geographical distribution:
\begin{equation}
\label{eq-se-context}
    \overline{\mathbf{C}}_{i,j}^{(t)} = \mathrm{Pooling}\big(\{\mathbf{G}^{(t)}\mathbf{d}_{j,k} \odot \widetilde{\mathbf{h}}_{k}^{(t)},v_k \in \mathcal{N}_K(v_i,v_j)\}\big),
\end{equation}
where $\mathbf{G}^{(t)}$ is the matrix for distance transformation, $\mathbf{d}_{j,k}$ denotes the internal spatial embedding between two neighbors $v_j$ and $v_k$, $\odot$ is the Hadamard product for element-wise multiplication. In practice, we utilize the mean pooling with the all-round distance integration to enable the model to comprehend how far other neighbors in $\mathcal{N}_K$ away from the interacted pair $v_j \rightarrow v_i$. As a result, every message from the neighbor $v_j$ to $v_i$ can carry the evolving context in \hide{multi-temporal set }$\mathcal{N}_K$ together with the spatial context $(\mathbf{d}_{i,j},\mathbf{d}_{j,i_1},...,\mathbf{d}_{j,i_K})$, rather than only the partial pairwise $\mathbf{d}_{i,j}$.

Finally, we present the contextual propagation module to combine each local neighboring location with the spatially evolving context. This procedure both considers a series of dynamic graph structures and the detailed contexts, which could be formulated in an interactive manner between adjacent time segments:

\begin{equation}
\label{eq-dy-conv}
    \mathbf{h}_{i,inter}^{(t)} =  \sum_{\tau \in [T_{t_1},T_{t_2}]}\sum_{v_j \in \mathcal{N}^{(\tau)}_i} \mathbf{W}^{(t)}_{\mathrm{Prop}}\big[\big(\mathbf{G}^{(t)}\mathbf{d}_{i,j}\odot \widetilde{\mathbf{h}}_{j}^{(\tau)}\big) \ocat \overline{\mathbf{C}}_{i,j}^{(t)}\big],
\end{equation}
where $\mathbf{W}^{(t)}_{\mathrm{Prop}}$ is \pending{shared }learning matrix for contextual propagation.
}

\subsection{Spatially Evolving Self-Supervised Learning}
\label{sec-ssl}
With the proposed two components \sconv and \tconv, we finally obtain the multi-temporal location embeddings $\{\mathbf{h}_{i,inter}^{(t_1)}, ..., \mathbf{h}_{i,inter}^{(t_T)}\}$. In the following sections, we use symbols $\{\mathbf{z}_{i}^{(t_1)}, ..., \mathbf{z}_{i}^{(t_T)}\}$ to represent these time-specific embeddings for simplicity.

As depicted in Figure \ref{fig-model}(b), we intend to design two essential self-supervised learning tasks to deal with the issue of the \liu{sparse} relationship labels \liu{between} locations, which enhances the representation learning process of \gnn beyond spatially intra-time and dynamically inter-time structure modeling.
Although it has been proved that applying the graph \liu{S}elf-\liu{S}upervised \liu{L}earning (SSL) strategy is effective for the general issue of scarce labeled data \cite{velickovic2019deep, zhu2021graph, lee2022relational}, current SSL framework\liu{s} always \liu{fail} to capture the complicated evolving patterns among locations under the spatially distributed environment. To this end, the learning procedure \liu{component, i.e.,} \ssl\liu{,} is proposed for multi-temporal location relationships in a self-supervised manner, which contains the global spatial information maximum $\mathcal{L}_{global}$ with the additional local evolving constraint $\mathcal{L}_{local}$. As a whole, we have the joint\hide{ly local-global} learning objective: 
\begin{equation}
\label{eq-all-loss}
\mathcal{L}_{ssl}=\lambda_{1}\mathcal{L}_{global} + \lambda_{2}\mathcal{L}_{local}+||\Theta||_2^2,
\end{equation}
where $\lambda_{1}$ and $\lambda_{2}$ are the hyper-parameters to balance the contributions of \hide{two }local-global \hide{evolving }loss functions, $||\Theta||_2^2$ is the L2 regularization.
\subsubsection{\B{Global \hide{Cross-Segment }Spatial Information Maximum}}

Inspired by the success of \liu{D}eep \liu{G}raph \liu{I}nfomax (DGI) \cite{velickovic2019deep}, we first develop the spatial information maximum objective to capture the global evolving patterns with preserving the gridding spatial dynamics. For time-evolving location graphs, the latent semantics of individual nodes do not stay unchanged over time segments. It is reasonable to treat the relational evolution as a generally smooth process since the human behaviors are gradually varying between adjacent times, indicating that the global surroundings around locations remain \licom{partially similar and complementary} at the next time segment. 

In practical scenarios, the location graph of a city can be partitioned into multiple urban grids \cite{li2020competitive}. Each urban grid $u$ gathers a cluster of spatially correlated location nodes $\{v_i|v_i \in u\}$ and provides the global surrounding information. All nodes located in an urban grid tend to share the similar spatial environment, which motivates us to introduce such knowledge into the model training. 

Different from DGI designed for a single graph, we perform the contrastive learning across the successive time-varying graph structures to fuse the global evolving information into the location representation. Specifically, we first utilize the grid-level pooling function to summarize the location representations to obtain the gridding vector $\mathbf{s}_u^{(t)} = \frac{1}{N_u}\sum_{v_i \in u}\mathbf{z}^{(t)}_i$, where $N_u$ is the number of locations in \hide{urban }grid $u$. After that, the goal is to maximize the cross-time mutual information between location-level representations and urban grid-level representations with the following loss function:
\begin{equation}
\begin{split}
\label{eq-global-loss}
    \mathcal{L}_{global}= \sum_{t\in\mathbb{T}}\Big(\sum_{v_i\in\mathcal{V}}\big(\mathbb{E}_{pos}[\mathrm{log}\mathcal{D}(\mathbf{z}_i^{(t)},\mathbf{s}^{(t-1)}_{u(i)})] \\ 
    + \frac{1}{|\mathrm{S}_i|} \sum_{v_j\in\mathrm{S}_i}\mathbb{E}_{neg}[\mathrm{log}(1-\mathcal{D}(\mathbf{z}_j^{(t)},\mathbf{s}^{(t-1)}_{u(i)}))]\big)\Big),
\end{split}
\end{equation}
where $u(i)$ denotes the urban grid where the location $v_i$ belongs to, the bilinear function $\mathcal{D}(\cdot, \cdot)$ is the discriminator to calculate the probability scores which estimates whether $v_i$ is located in the grid $u$. Note that the natural positive sample $(\mathbf{z}_i^{(t)},\mathbf{s}^{(t-1)}_{u(i)})$ is extracted from temporal views, while the negative pairs are generated from the gridding sampler $\mathrm{S}_i$. In particular, the location $v_i$ at the $t$-th time segment and the corresponding urban grid $u(i)$ at the last segment $t-1$ are regarded as a positive pair. The heuristic sampler aims to distinguish effective negative pairs based on geographical information rather than random sampling.
\begin{equation}
\label{eq-neg-sampler}
    \mathrm{S}_i = \{v_j|d_1<ManhattanDist\big[u(i),u(j)\big]<d_2\},
\end{equation}
where the function $ManhattanDist[\cdot, \cdot]$ over the gridding city map returns the Manhattan distance between the two urban grids $u(i)$ and $u(j)$. For the sake of avoiding inadequate locations which are \hide{too close to $u(i)$ or too far away from $u(i)$}\pending{too close or too far away}, we use $d_1$ and $d_2$ to define the appropriate sampling scope of spatial areas (we empirically set $d_1$ and $d_2$ as 2 and 6 in practice) for high-quality negative samples generation. 

\subsubsection{\B{Local Relational Evolving Constraint}}
Besides the global evolving correlations, the relevance among time-specific relationships \liu{is} also evolutionary from a local perspective. We further introduce the edge-level relational evolving constraint to complement the global grid-level evolution in \ssl. The key idea of this constraint is to explore if the relation \hide{$r_{ij}^{(t)}$}\pending{$r_{ij}\in \mathcal{R}$} between $v_i$ and $v_j$ still remains \pending{at time $t$\hide{(i.e., $r_{ij}^{(t)}$)}} when \hide{a relation}\pending{this relation}\hide{$(v_i, r_{ij}^{(t-1)}, v_j)$}\hide{$(v_i,v_j,r_{ij},t-1)$} exists at the last $t-1$ time segment. We observe some specific short-term relations would disappear at the subsequent segment $t$, while a bundle of long-term influential relations would continue to survive. Therefore, considering the complex evolving pattern is of great importance due to the potential time continuity for location relationships. We first acquire the relational edge representation $\mathbf{e}_{ij}^{(t)} = \mathbf{h}_{i}^{(t)}\odot\mathbf{h}_{j}^{(t)}$. To further explicitly preserve the relation-aware evolving patterns between adjacent \hide{segments}\pending{times}, we define the following local objective:
\begin{equation}
\small
\begin{split}
\label{eq-local-loss}
    \mathcal{L}_{local}= \sum_{t\in\mathbb{T}}\sum_{e_{ij}\in\mathcal{E}^{(t)}} \Big(\delta(r_{ij}^{(t)},r_{ij}^{(t-1)})\mathrm{log}[\varphi(\mathbf{e}_{ij}^{(t)},\mathbf{e}_{ij}^{(t-1)})] \\
    + \big(1-\delta(r_{ij}^{(t)},r_{ij}^{(t-1)})\big)\mathrm{log}[1-\varphi(\mathbf{e}_{ij}^{(t)},\mathbf{e}_{ij}^{(t-1)})]\Big), 
\end{split}
\end{equation}
where $\mathcal{E}^{(t)}$ is the relational edge set at the $t$-th time segment, the Kronecker delta function $\delta(\cdot,\cdot)$ outputs 1 only if the relationship $r_{ij}$ remains the same from time $t-1$ to $t$, the MLP function $\varphi(\cdot,\cdot)$ is adopted to calculate the evolving probability.
\vspace{-1ex}
\subsection{Time-aware Relationship Inference\hide{ and Optimization}}
After training the proposed \gnn through the spatially evolving self-supervised learning stage, we then take advantage of the well-trained model to predict the time-aware location relationships. Given a pair of locations $(v_i,v_j)$, the model can learn the pairwise multi-slot location embeddings $\{(\mathbf{z}^{(t)}_i,\mathbf{z}^{(t)}_j)|t\in\mathbb{T}\}$ over the whole time segments $\mathbb{T}$. Finally, we adopt the time-specific DistMult factorization \cite{yang2015embedding} as the scoring function for prediction at each time.
\begin{equation}
\small
\label{eq-pred}
    \hat{y}^{(t)}_{ij,r} = \sigma\big(\mathbf{z}^{(t)^{T}}_i\mathbf{W}^{(t)}_r\mathbf{z}^{(t)}_j\big),
\end{equation}
where the time-aware diagonal matrix $\mathbf{W}^{(t)}_r$ is the learnable parameter for relationship $r$, $\sigma$ stands for the sigmoid function. Then the cross entropy loss function between the predicted probability $\hat{y}^{(t)}_{ij,r}$ and the label $y^{(t)}_{ij,r}$ is used to jointly optimize the model under a time aggregated multi-task learning manner:
\begin{equation}
\label{eq-rel-loss}
\small
    \mathcal{L}_{rel}= \sum_{t\in\mathbb{T}}\sum_{(v_i,r,v_j)\in\mathcal{Y}_{trn}}\Big(y^{(t)}_{ij,r}\mathrm{log}\hat{y}^{(t)}_{ij,r}+(1-y^{(t)}_{ij,r})\mathrm{log}(1-\hat{y}^{(t)}_{ij,r})\Big), 
\end{equation}
where $\mathcal{Y}_{trn}$ is the training edge set, $y^{(t)}_{ij,r}$ indicates whether there exists the relationship $r$ between $v_i$ and $v_j$ at the time segment $t$.

\section{Experiments} \label{sec-exp}

\hide{In this section, we first introduce the constructed datasets, and then conduct extensive experiments to comprehensively evaluate the proposed model compared against the state-of-the-art methods.}
\licom{
In this section, we conduct extensive experiments to evaluate the proposed \model compared against the state-of-the-art methods. The code of \model is available at \url{https://github.com/PaddlePaddle/PaddleSpatial/tree/main/research/SEENet}.
}
\vspace{-1.4ex}

\begin{table*}[t]
	\renewcommand{\arraystretch}{0.94}
        \caption{Overall performance\hide{ comparison} on Business-RD (\textsc{Beijing} and \textsc{Tokyo}) and Mobi-RD (\textsc{New York} and \textsc{Chicago}). We conduct experiments with five random seeds and \hide{the results are reported as mean$\pm$std.}\pending{report the average performance together with the standard deviation.} \hide{We use the paired t-test with a significance level at 0.05 on all reported results.}}
        \vspace{-2.5ex}
        \label{table-main-exp}
	\centering
	\scalebox{.9}{
	\begin{tabular}{c|cc|cc|cc|cc}
		\toprule
				\multirow{2}{*}{Method} & \multicolumn{2}{c|}{\textsc{Beijing}} & \multicolumn{2}{c|}{\textsc{Tokyo}} & \multicolumn{2}{c|}{\textsc{New York}} & \multicolumn{2}{c}{\textsc{Chicago}} \\
		\cmidrule{2-9}
        & MRR@10 & HR@10 & MRR@10 & HR@10 & MRR@10 & HR@10 & MRR@10 & HR@10 \\ 
         \midrule 
    
         GCN & 0.1278{\small{$\pm$0.005}} & 0.2873{\small{$\pm$0.010}} & 0.1386{\small{$\pm$0.003}} & 0.3034{\small{$\pm$0.012}} & 0.1213{\small{$\pm$0.005}} & 0.3184{\small{$\pm$0.012}} & 0.1052{\small{$\pm$0.002}} & 0.3467{\small{$\pm$0.008}}  \\  
         PathGCN & 0.1311{\small{$\pm$0.006}} & 0.3191{\small{$\pm$0.015}} & 0.1380{\small{$\pm$0.004}} & 0.3387{\small{$\pm$0.014}} & 0.1375{\small{$\pm$0.005}} & 0.3478{\small{$\pm$0.012}} & 0.1056{\small{$\pm$0.006}} & 0.3471{\small{$\pm$0.025}} \\
         CompGCN & 0.1637{\small{$\pm$0.001}} & 0.4482{\small{$\pm$0.005}} & 0.1394{\small{$\pm$0.004}} & 0.3590{\small{$\pm$0.005}} & 0.1423{\small{$\pm$0.012}} & 0.3772{\small{$\pm$0.021}} & 0.0923{\small{$\pm$0.002}} & 0.3305{\small{$\pm$0.002}} \\
		 MixHop & 0.1703{\small{$\pm$0.001}} & 0.4582{\small{$\pm$0.006}} & 0.1412{\small{$\pm$0.003}} & 0.3618{\small{$\pm$0.006}} & 0.1488{\small{$\pm$0.014}} & 0.3528{\small{$\pm$0.026}} & 0.1094{\small{$\pm$0.006}} & 0.3491{\small{$\pm$0.011}} \\
          NL-GNN & 0.1811{\small{$\pm$0.004}} & 0.4233{\small{$\pm$0.009}} & 0.1639{\small{$\pm$0.003}} & 0.4134{\small{$\pm$0.009}} & 0.1785{\small{$\pm$0.006}} & 0.4021{\small{$\pm$0.009}} & 0.1107{\small{$\pm$0.001}} & 0.3695{\small{$\pm$0.008}}\\
          
	 \midrule 
         GCA & 0.1561{\small{$\pm$0.016}} & 0.3416{\small{$\pm$0.035}} & 0.1650{\small{$\pm$0.003}} & 0.4010{\small{$\pm$0.012}} & 0.1188{\small{$\pm$0.002}} & 0.3449{\small{$\pm$0.006}} & 0.1116{\small{$\pm$0.002}} & 0.3736{\small{$\pm$0.005}}  \\  
         DGI & 0.1776{\small{$\pm$0.007}} & 0.3893{\small{$\pm$0.012}} & 0.1738{\small{$\pm$0.001}} & 0.3976{\small{$\pm$0.004}} & 0.1538{\small{$\pm$0.006}} & 0.3985{\small{$\pm$0.023}} & 0.1098{\small{$\pm$0.002}} & 0.3587{\small{$\pm$0.010}}\\
         RGRL & 0.1952{\small{$\pm$0.006}} & 0.4216{\small{$\pm$0.015}} & \underline{0.1775{\small{$\pm$0.007}}} & \underline{0.4253{\small{$\pm$0.016}}} & 0.1624{\small{$\pm$0.001}} & 0.3939{\small{$\pm$0.006}} & 0.1107{\small{$\pm$0.002}} & 0.3676{\small{$\pm$0.005}}  \\
    	 \midrule 
         EvolveGCN & 0.2123{\small{$\pm$0.003}} & 0.4870{\small{$\pm$0.001}} & 0.1634{\small{$\pm$0.003}} & 0.4070{\small{$\pm$0.010}} & 0.2054{\small{$\pm$0.002}} & 0.4518{\small{$\pm$0.010}} & 0.0976{\small{$\pm$0.001}} & 0.3290{\small{$\pm$0.001}} \\  
          ROLAND & 0.2127{\small{$\pm$0.013}} & 0.4966{\small{$\pm$0.030}} & 0.1607{\small{$\pm$0.004}} & 0.4133{\small{$\pm$0.006}} & 0.1980{\small{$\pm$0.004}} & 0.4571{\small{$\pm$0.009}} & \underline{0.1237{\small{$\pm$0.002}}} & \underline{0.3859{\small{$\pm$0.008}}} \\
         \midrule 
         DecGCN & 0.1758{\small{$\pm$0.001}} & 0.4175{\small{$\pm$0.002}} & 0.1552{\small{$\pm$0.000}} & 0.3686{\small{$\pm$0.007}} & 0.1700{\small{$\pm$0.002}} & 0.4142{\small{$\pm$0.005}} & 0.1112{\small{$\pm$0.002}} & 0.3466{\small{$\pm$0.001}}  \\
         IRGNN & 0.1807{\small{$\pm$0.012}} & 0.4176{\small{$\pm$0.025}} & 0.1299{\small{$\pm$0.001}} & 0.3136{\small{$\pm$0.010}} & 0.1638{\small{$\pm$0.005}} & 0.3843{\small{$\pm$0.006}} & 0.1123{\small{$\pm$0.003}} & 0.3408{\small{$\pm$0.005}} \\
         DeepR & \underline{0.2184{\small{$\pm$0.002}}} & \underline{0.5257{\small{$\pm$0.006}}} & 0.1662{\small{$\pm$0.001}} & 0.3902{\small{$\pm$0.005}} & 0.1988{\small{$\pm$0.001}} & 0.4496{\small{$\pm$0.003}} & 0.1058{\small{$\pm$0.004}} & 0.3628{\small{$\pm$0.010}}  \\  
         PRIM & 0.1973{\small{$\pm$0.001}} & 0.4992{\small{$\pm$0.002}} & 0.1454{\small{$\pm$0.003}} & 0.3990{\small{$\pm$0.006}} & \underline{0.2229{\small{$\pm$0.007}}} & \underline{0.5008{\small{$\pm$0.002}}} & 0.1021{\small{$\pm$0.001}} & 0.3634{\small{$\pm$0.003}} \\
         
       \midrule 
	\model & \B{0.2545{\small{$\pm$0.003}}} & \B{0.5524{\small{$\pm$0.007}}} & \B{0.2314{\small{$\pm$0.003}}} & \B{0.4880{\small{$\pm$0.009}}} & \B{0.2526{\small{$\pm$0.003}}} & \B{0.5376{\small{$\pm$0.009}}} & \B{0.1506{\small{$\pm$0.002}}} & \B{0.4338{\small{$\pm$0.012}}}
		\\ \bottomrule 
		
	\end{tabular}}
	\vspace{-2.5ex}
\end{table*}

\begin{table}
\small
	\centering
    \caption{Statistics of four real-world datasets.}
    \vspace{-4mm}
	\scalebox{0.9}{\begin{tabular}{ccccccc}
		\toprule
		Dataset	& Beijing & Tokyo& New York & Chicago \\
		\midrule
            Relation Type & Business & Business & Mobility & Mobility \\
            Relation Source & Map Query & Check-in & By Taxi & By Bike \\
            \# Nodes & 30,114 & 3,013 & 1,587 & 483 \\
            \# \pending{Relations at $t_1$}\hide{Morning Edges} & 3,270  & 1,820 & 647 & 1,059 \\
            \# \pending{Relations at $t_2$}\hide{Midday Edges} & 96,233  & 3,991 & 7,016 & 5,894 \\
            \# \pending{Relations at $t_3$}\hide{Night Edges} & 97,829  & 9,419 & 7,909 & 6,275 \\
            \# \pending{Relations at $t_4$}\hide{Midnight Edges} & 4,155  & 1,446 & 4,523 & 1,322 \\
            \hide{\pending{Graph Sparsity} & 99.98\% & 99.82\% & 99.20\% & 93.76\% \\}
		\bottomrule
	\end{tabular}}
	\vspace{-5.2mm}
	\label{table-dataset}
\end{table}
\subsection{Experiment Settings}
\subsubsection{\B{Datasets}} Our experiments are conducted on four real-world citywide datasets from two distinct relationship learning domains. The first two urban \textbf{business-based} \hide{relationship }datasets (\textit{Beijing} and \textit{Tokyo}) are derived from commercial behaviors of users, while the other two (\textit{New York} and \textit{Chicago}) are urban \textbf{mobility-based} \hide{relationship}datasets constructed from \hide{traffic}trajectory data to ensure the data diversity.\hide{which ensures data diversity.} 
\begin{itemize}[leftmargin=*,topsep=2pt]
    \item \B{Business-based Relational Data.} (Business-RD for short) In the scene of urban business, it has been studied that there are two significant behavior-driven relationships among locations \cite{li2020competitive,chen2021points}, i.e., \textit{competitive} and \textit{complementary} relationships, which could be generated\hide{reflected} from session-based location query data or check-in data. \hide{Different types of user behaviors conveys different relational semantics. }Therefore, we follow previous works \cite{li2020competitive,chen2021points,liu2020decoupled} to construct \licom{the Business-RD (\textbf{Beijing} and \textbf{Tokyo}) from query dataset QueryBJ and check-in dataset Foursquare \cite{yang2014modeling} respectively.} \hide{We first use the QueryBJ dataset collected from Baidu Maps \footnote{https://map.baidu.com}, which contains millions of map search query logs\hide{ from 1/1/2019 to 1/9/2019} in \textbf{Beijing}. Each query log provides the user's query session with timestamps. Following the works \cite{chen2021points,liu2020decoupled}, the two types of relations between POIs $v_i$ and $v_j$ are defined as: 
    
    \underline{(1)} Users viewed $v_i$ also viewed $v_j$ within a query session\hide{ (\textit{competitive}}; 
    
    \underline{(2)} Users viewed $v_i$ then viewed $v_j$ across different \hide{query }sessions\hide{(\textit{complementary})}.
    
    As for the public Foursquare dataset \cite{yang2014modeling}, we collect the check-in data from the representative city \textbf{Tokyo}. All visited POIs of a user are classified into several categorical trips (e.g., all daily restaurant POIs form a categorical trip). Similarly, inspired by the work \cite{li2020competitive}, we have the two kinds of relationships as follows:
    
    \underline{(3)} Users visited $v_i$ also visited $v_j$ within a categorical trip\hide{ (\textit{competitive})};
    
    \underline{(4)} Users visited $v_i$ then visited $v_j$ across different categories\hide{ categorical trips}\hide{(\textit{complementary}}.  

    According to these \pending{relationship} studies \cite{mcauley2015inferring, li2020competitive, chen2021points}, we refer to (1) and (3) as \textit{competitive} relationships while the other (2) and (4) are \hide{considered}known as \textit{complementary} relationships.}
    \item \B{Mobility-based Relational Data.} (Mobi-RD for short) Since urban mobility is another important aspect to reflect the dynamic location correlations, we further extend the relationship in the \pending{mobile} trajectory domain\hide{, which ensures data diversity}. \licom{In specific, we utilize the taxi and bike trajectory datasets NYCTaxi \footnote{https://nyc.gov/site/tlc/about/tlc-trip-record-data.page} and DivvyBike\footnote{https://ride.divvybikes.com/system-data} to generate the Mobi-RD (\textbf{New York} and \textbf{Chicago}), which includes \textit{high-flow} and \textit{low-flow} relationships according to the flowing degree.}\hide{ For the taxi driving data, the public NYCTaxi \footnote{https://nyc.gov/site/tlc/about/tlc-trip-record-data.page} includes the order records traveling throughout \textbf{New York} City, while the opening dataset DivvyBike\footnote{https://ride.divvybikes.com/system-data} collects bike riding orders from the bike sharing system Divvy\hide{of people daily using} in \textbf{Chicago}. In this scene, the POI is specified as a bike station or an urban region. Each trajectory by taxi or bike includes the pick-up and drop-off locations with timestamps, which connects a pair of POIs\hide{ $(v_i,v_j)$}. After counting the overall records for flow-based relevancy\hide{ $fr_{ij}$}, we label the two meaningful relationships between POI pairs, namely \textit{high-flow} (top 25\% high) and \textit{low-flow} (top 50\% high) relationships according to the flowing degree.}
\end{itemize}
\hide{\noindent}\hide{\hide{Fine-grained dynamic POI relationship decomposition.}Multi-time relationship decomposition.} As aforementioned, the special location relationships are influenced by time\hide{can be time-varying} in reality since the above user behaviors and trajectories are dependent on different time periods in a day (e.g., two restaurants tend to be competitive at midday instead of midnight) \cite{yuan2013time}. Thus, we reasonably detail the \hide{business-based and flow-based }relationships with time-aware refinements. In practice, as suggested by the literature \cite{li2017time} with considering the \hide{expert knowledge}\pending{real-life experience} and \pending{data analysis}\hide{the \pending{data analysis in}\hide{observation from} Figure \ref{fig-analysis}}, we evenly split a day into four segments $t_1$\textasciitilde$t_4$, i.e., \textit{morning}, \textit{midday}, \textit{night}, and \textit{midnight}. We construct the above four datasets at each time \hide{segment }according to \hide{the data}\pending{timestamps}, and finally obtain the \pending{\hide{time-aware}\licom{\B{multi-temporal Business-RD and Mobi-RD} in Table \ref{table-dataset}}}. \licom{The details of relationship construction \hide{with processing codes }are included in Appendix \ref{A-dataset}.}

\vspace{-0.8ex}
\subsubsection{\B{Setup}}
\licom{
Following the previous relationship-based works \cite{li2020competitive,liu2021item}, we randomly sample 10\% of relational edges for testing and 10\% of edges as the validation set at each time segment, while the remaining 80\% of relations are utilized to construct the dynamic location graph for training. \hide{\pending{Meanwhile,}\hide{Note that} we guarantee that the testing and validation sets $\mathcal{Y}_{pred}$ are absolutely independent of the training set $\mathcal{Y}_{trn}$ across all time segments, meaning that every relational edge in $\mathcal{Y}_{pred}$ at a certain time will not appear in $\mathcal{Y}_{trn}$ at any time segment.}\hide{once an edge between $v_i$ and $v_j$ appears in the testing or validation set at time $t$, we \hide{request}\pending{limit} not appearing in the training graph.}We also replace the destination node of each edge with other random locations for negative sampling.}
\vspace{-0.8ex}
\hide{
\subsubsection{\B{Evaluation Metrics}}\hide{ To comprehensively evaluate the performance of our model and baselines in discovering dynamic POI relationships,} We conduct time-specific link prediction experiments \pending{to rank candidate POIs} at each time. \hide{For each testing edge, the task aims to rank candidate POIs of all time-aware relationships for the source POI node. ranking performance}As introduced in \cite{liu2020decoupled} for relationship inference, we adopt widely-used Mean Reciprocal Ranking (\textit{MRR@k}) and Hit Rate (\textit{HR@k}) \hide{in top \textit{k} list }as evaluation metrics.\hide{ The details are introduced in Appendix \ref{A-metric}.}

\hide{which mathematically are defined as:
\hide{The mathematical definitions of \textit{MRR@k} and \textit{HR@k} are given:}
\begin{align}
\label{eq-metric}
    \pending{MRR@k} &= \frac{1}{M}\sum_{i=1}^M \mathcal{I}_{rank_i\le k}\cdot\frac{1}{rank_i}\\
   \hide{MRR@k &= \frac{1}{k}\sum_{i=1}^k \mathcal{I}_{rank_i\le k}\cdot\frac{1}{i}\\}
    HR@k &= \frac{1}{M}\sum_{i=1}^M \mathcal{I}_{rank_i\le k}
\end{align}
where $M$ is the size of testing set, $rank_i$ represents the ranking index of the POI $v_i$, $\mathcal{I}$ is the indicator function ($\mathcal{I}_{rank_i\le k}=1$ iff $v_i$ is among the top $k$ and 0 others).}
}
\vspace{-0mm}
\subsubsection{\B{Baselines and Evaluation Metrics}}
We compare\hide{ the proposed} \model with a variety of advanced GNN methods for dynamic location relationship inference: (i) The relational-based GNNs (\B{GCN} \cite{kipf2017semi}, \B{PathGCN} \cite{eliasof2022pathgcn}, \B{CompGCN} \cite{vashishth2020composition}, \B{MixHop} \cite{abu2019mixhop}, and \B{NL-GNN} \cite{liu2021non}) are typical graph structure learning models considering node correlations and contexts. (ii, iii) We also select recent graph self-supervised learning models (\B{DGI} \cite{velickovic2019deep}, \B{GCA} \cite{zhu2021graph}, and \B{RGRL} \cite{lee2022relational}) and \pending{\hide{recent }snapshot-based} dynamic GNNs for link prediction (\B{EvolveGCN} \cite{pareja2020evolvegcn} and \B{ROLAND} \cite{you2022roland}). (iv) Moreover, \model is compared with state-of-the-art relationship prediction methods (\B{DecGCN} \cite{liu2020decoupled}, \B{IRGNN} \cite{liu2021item}, \B{DeepR} \cite{li2020competitive}, and \B{PRIM} \cite{chen2021points}) which are designed for learning relational and \hide{geographical}\licom{spatial} dependencies. \licom{We conduct time-specific link prediction experiments \pending{to rank candidate locations} at each time. \hide{For each testing edge, the task aims to rank candidate POIs of all time-aware relationships for the source POI node. ranking performance}As introduced in \cite{liu2020decoupled} for relationship inference, we adopt \hide{widely-used }Mean Reciprocal Ranking (\textit{MRR@k}) and Hit Rate (\textit{HR@k}) \hide{in top \textit{k} list }as evaluation metrics.} \hide{More details of baselines can be found in Appendix \ref{A-baseline}}\licom{The \zhou{baseline} descriptions, parameter settings, and more experimental details are introduced in Appendix \ref{A-detail}.}

\vspace{-2ex}

\subsection{Overall Performance Comparison}
\hide{\subsubsection{Overall Comparison}}We first compare the overall experimental results on four real-world relationship datasets, where the evaluation metrics are calculated over all \pending{time periods}\hide{ segments} to reflect the general prediction performance\hide{ (All detailed multi-temporal results are provided in Appendix \ref{A-main-exp})}. As presented in Table \ref{table-main-exp}, we report the metrics MRR@10 and HR@10 in the relationship prediction results following the previous work \cite{liu2020decoupled}. The \textbf{values in boldface} indicate the best results, while the \underline{underlined values} signify the second-best results. On the whole, it is observed that our \model consistently outperforms all different baseline methods by an obvious margin on each dataset. \pending{In specific, compared with the second-best model on four datasets, \model improves the MRR@10 by 16.5\%, 30.4\%, 13.3\%, and 21.7\% \hide{,and improves the HR@10 by 5.1\%, 14.7\%, 7.3\%, and 12.4\% }respectively. }We further have the following observations and findings.

As we can see, the relational-based GNNs roughly perform worse than other types of baselines since they only leverage the graph topological structures with learning node relations. It is not surprising that the recently proposed PathGCN just achieves comparable results with GCN because the learned spatial operators with random paths are uncertain and sometimes may go against the relationship prediction. Moreover, CompGCN performs slightly better\hide{ than GCN} because of the ability to distinguish multiple relations, while the non-local message-passing models (Mixhop and NL-GNN) can explicitly learn multi-hop relational contexts and further improve the performance. In general, the above GNNs are not ideal without spatial and dynamic learning schemes, which verifies that simply aggregating relational edges is not adequate for complex dynamic location graphs. As to self-supervised learning methods, RGRL outperforms the other two baselines, as it can enable \liu{S}elf-\liu{S}upervised \liu{L}earning (SSL)\hide{ on sparse POI graphs} with capturing the augmentation-invariant relationship at the same time. Note that although both DGI and RGRL adopt the global-local SSL architecture on graphs, our \model incorporates the spatial distribution and evolving pattern into SSL beyond the traditional framework and performs much better. 

From the perspective of dynamic relationship modeling, since EvolveGCN and ROLAND can take advantage of dynamic propagation cross times to preserve time-aware relational characteristics, these models exhibit considerable improvement over classic GNNs. Furthermore, we can see that the latest ROLAND can not always outperform EvolveGCN due to the failure of recurrent designs on some datasets, indicating dynamic GNNs are still \hide{insufficiently powerful}\pending{not powerful enough} in the domain of location graph learning. From the perspective of special relationship prediction approaches, we notice that spatial-oriented models (DeepR and PRIM) tend to perform better than DecGCN and IRGNN as a result of considering essential geographical information and adapting the relationship learning \licom{in the scenario of location graphs}. One exception is on the Chicago dataset \pending{with \liu{the} fewest relations}\hide{, where the \hide{bike-based flow graph} is quite small}. \hide{In this case, }The potential reason is that DeepR and PRIM requiring enough neighbors lose their effectiveness when learning spatial contexts on small graphs. By contrast, our model can fully capture both spatial correlations and evolving dynamics with graph convolution-level and \pending{SSL-level} enhancements. Therefore, \model is much more effective for multi-temporal relationship inference among locations.

\hide{
\subsection{Time-specific Performance Analysis}
\hide{Then we \hide{perform}conduct \liu{a} more detailed analysis for time-specific performance compared with several most competitive baselines. Results of all baselines on each dataset are included in Appendix \ref{A-main-exp}.}

Table \ref{table-time-exp} exhibits the multi-temporal relationship prediction results over all times in a day. The proposed \model provides stable performance gains over \revise{major competitive} baselines at all time periods, which consistently verifies the superiority of our multi-slot spatial learning framework in time-aware predictions. We also observe that most baselines perform worse at midnight due to the sparsity of location relationships. We \liu{observe} that the graph at midnight with fewer people activities only contains limited environmental information for location relationship inference. Moreover, we \liu{find} that dynamic GNNs (EvolveGCN and ROLAND) can partly alleviate the above issue thanks to the capability of involving the relational knowledge from the previous time, which agrees with the fact that considering dynamic relationships is essential and valuable. However, the \hide{time-specific }performance is still unsatisfactory. By contrast, our model can achieve stable and accurate prediction results at any time, demonstrating the exhaustive effectiveness of \model.

\begin{table}
	\centering
    \caption{Time-specific performance evaluation (MRR@10) over \hide{segments }times on two datasets of Business-RD and Mobi-RD.}
    \label{table-time-exp}
    \vspace{-4mm}
	\scalebox{0.85}{\begin{tabular}{c|c|cccc}
		\toprule
		  \B{Dataset} & \B{Method} & \B{Morning} & \B{Midday} & \B{Night} & \B{Midnight} \\
		\midrule
            \multirow{7}{*}{\shortstack{\textsc{Beijing}}} 
            & NL-GNN & 0.2302 & 0.1803 & 0.1821 & 0.1340 \\
            & RGRL & 0.2232 & 0.1901 & 0.2018 & 0.1352\\
            & EvolveGCN & \underline{0.2718} & 0.2072 & 0.2169 & \underline{0.1760} \\
            & ROLAND & 0.2402 & 0.2080 & \underline{0.2187} & 0.1567 \\
            & DeepR & 0.2152 & \underline{0.2247} & 0.2159 & 0.1334 \\
            & PRIM & 0.2114 & 0.2005 & 0.1962 & 0.1409 \\
            & \model & \B{0.3206} & \B{0.2442} & \B{0.2621} & \B{0.2602} \\
            \midrule
            \multirow{7}{*}{\shortstack{\textsc{New York}}} 
            & NL-GNN & 0.1386 & 0.2042 & 0.2029 & 0.1532 \\
            & RGRL & 0.1778 & 0.1419 & 0.1607 & 0.1850\\
            & EvolveGCN & 0.1463 & 0.2185 & 0.2259 & 0.1576 \\
            & ROLAND & 0.1799 & 0.2027 & 0.2129 & \underline{0.1993} \\
            & DeepR & 0.1838 & 0.2122 & 0.2080 & 0.1734 \\
            & PRIM & \underline{0.2072} & \underline{0.2276} & \underline{0.2420} & 0.1976 \\
            & \model & \B{0.2722} & \B{0.2367} & \B{0.2714} & \B{0.2415} \\
		\bottomrule
	\end{tabular}}
	\vspace{-5mm}
\end{table}
}

\begin{table}
	\centering
    \caption{Ablation studies with the metric MRR@10\hide{for spatial and dynamic designs}.}
    \label{table-abla-exp}
    \vspace{-4mm}
	\scalebox{0.85}{\begin{tabular}{c|cccc}
		\toprule
		   \B{Variants} & \B{Beijing} & \B{Tokyo} & \B{New York} & \B{Chicago} \\
		\midrule
            \textsf{\model-SEC} & 0.1780 & 0.1636 & 0.1548 & 0.1218 \\
            \textsf{\model-RS} & 0.1792 & 0.1901 & 0.1744 & 0.1323 \\
            \textsf{\model-LD} & 0.2149 & 0.169 & 0.1951 & 0.1193 \\
            \textsf{\model-C} & 0.2469 & 0.1718 & 0.2169 & 0.1215 \\
            \midrule
            \textsf{\model-SSL} & 0.2321 & 0.2042 & 0.1937 & 0.1371 \\
            \textsf{\model-L} & 0.2466 & 0.2157 & 0.2116 & 0.1441 \\
            \textsf{\model-G} & 0.2463 & 0.2215 & 0.2493 & 0.1429 \\
            \midrule
            \model & \B{0.2545} & \B{0.2314} & \B{0.2526} & \B{0.1506} \\
		\bottomrule
	\end{tabular}}
	\vspace{-5mm}
\end{table}
\vspace{-1mm}
\subsection{Impact of Spatial and Dynamic Designs}
\label{sec-abla}
\hide{In this section, we conduct the ablation study to evaluate the impact of key designs in the proposed \model, including model\hide{ structure} analysis for \gnn and self-supervised learning analysis for \ssl.}
\subsubsection{How \model Architecture Design Helps (Model Analysis)}
To investigate the contribution of each component in our designed multi-slot spatial graph convolutions \gnn, we compare \model with the following variants on four datasets \revise{in Table \ref{table-abla-exp}}.
\begin{itemize}[leftmargin=*,topsep=1pt]
    \item \revise{\B{\textsf{\model-SEC}} replace the whole \gnn with a classic GCN.}
    \item \B{\textsf{\model-RS}} removes spatial \sconv for intra-time learning.
    \item \B{\textsf{\model-LD}} removes dynamic \tconv for inter-time learning.
    \item \B{\textsf{\model-C}} drops the evolving context\hide{dynamic aggregation} of \tconv in Eq. (\ref{eq-dy-conv}).
\end{itemize}
\hide{The experimental results of MRR@10 are reported in Figure \ref{abla-exp-model-business} and \ref{abla-exp-model-flow}.}It is obvious that \revise{other variants of \model outperform \textsf{\model-SEC} and} the performance generally decreases when we gradually remove the graph learning components. In particular, if we replace the time-aware contextual aggregation with a simple GCN-style function when performing inter-time interactions, we find that \textsf{\model-C} get\liu{s} worse, \pending{which confirms that it is beneficial to integrate the \hide{spatial influence}\licom{spatially evolving context for \problem}\hide{ in the dynamic learning process}.} \hide{The results also show that other variants of \model outperform \textsf{\model-SEC}.} Furthermore, \textsf{\model-LD} performs better than \textsf{\model-RS} on Beijing and New York datasets while the converse is observed on the other two datasets, proving both spatial and dynamic modeling can play significant roles across different scenarios. In summary, the results highlight the importance of designing synergistic \model architecture to combine geographical factors and dynamic relationships.

\subsubsection{How \hide{Spatially Evolving }Self-Supervised Learning Design Helps (\ssl analysis)}
We also conduct essential experiments to validate the effectiveness of \hide{spatially evolving self-supervised learning }the well-designed \ssl with removing different objectives.
\begin{itemize}[leftmargin=*,topsep=1pt]
    \item \B{\textsf{\model-SSL}} w/o\hide{without} pre-training, i.e., removing the whole \ssl.
    \item \B{\textsf{\model-L}} w/o\hide{drops} the loss of local relational evolving constraint.
    \item \B{\textsf{\model-G}} w/o\hide{drops} the loss of global spatial information maximum.
\end{itemize}
As we can see \hide{in Figure \ref{abla-exp-ssl-business} and \ref{abla-exp-ssl-flow}}\revise{in Table \ref{table-abla-exp}}, there is a consistent performance degradation when excluding either global-view loss or local-view loss. \textsf{\model-SSL} performs even worse than \textsf{\model-G} and \textsf{\model-L} when dropping global and local learning objectives at once, showing both of them can contribute to model training. The observation verifies that considering global spatial distribution as well as local evolving relationship patterns in a self-supervised learning manner is critical for \hide{dynamic}location relationship inference.

\hide{
\begin{figure}
\setlength{\abovecaptionskip}{2.mm}
\setlength{\belowcaptionskip}{-0.cm}
  \centering
  \subfigure{
    \label{abla-exp-model-business} 
    \includegraphics[width=0.43\columnwidth]{figure/ablation/abla-model-business-mrr-caption-new1.pdf}}
      \subfigure{
    \label{abla-exp-model-flow}
    \includegraphics[width=0.43\columnwidth]{figure/ablation/abla-model-mobi-mrr-caption-new1.pdf}}
\\[-3ex]
  \subfigure{
    \label{abla-exp-ssl-business} 
    \includegraphics[width=0.43\columnwidth]{figure/ablation/abla-ssl-business-mrr-caption-blue-new.pdf}}
    \subfigure{
    \label{abla-exp-ssl-flow} 
    \includegraphics[width=0.43\columnwidth]{figure/ablation/abla-ssl-mobi-mrr-caption-blue-new.pdf}}
  \vspace{-3mm}
  \caption{Contribution of spatial and dynamic designs.\hide{ in our proposed \model on four citywide datasets.}}
  \vspace{-4mm}
  \label{fig-ablation}
\end{figure}
}
\vspace{1ex}
\hide{\subsection{Parameter Sensitive Analysis}}

 We further explore the influence of various important parameters in \model. More experimental results are in Appendix \ref{A-para-exp}.
\vspace{-1ex}
\subsubsection{Coefficients of global and local loss functions}
\begin{figure}
\setlength{\abovecaptionskip}{2.mm}
\setlength{\belowcaptionskip}{-0.cm}
  \centering
  \subfigure{
    \label{fig-para-global} 
    \includegraphics[width=0.31\columnwidth]{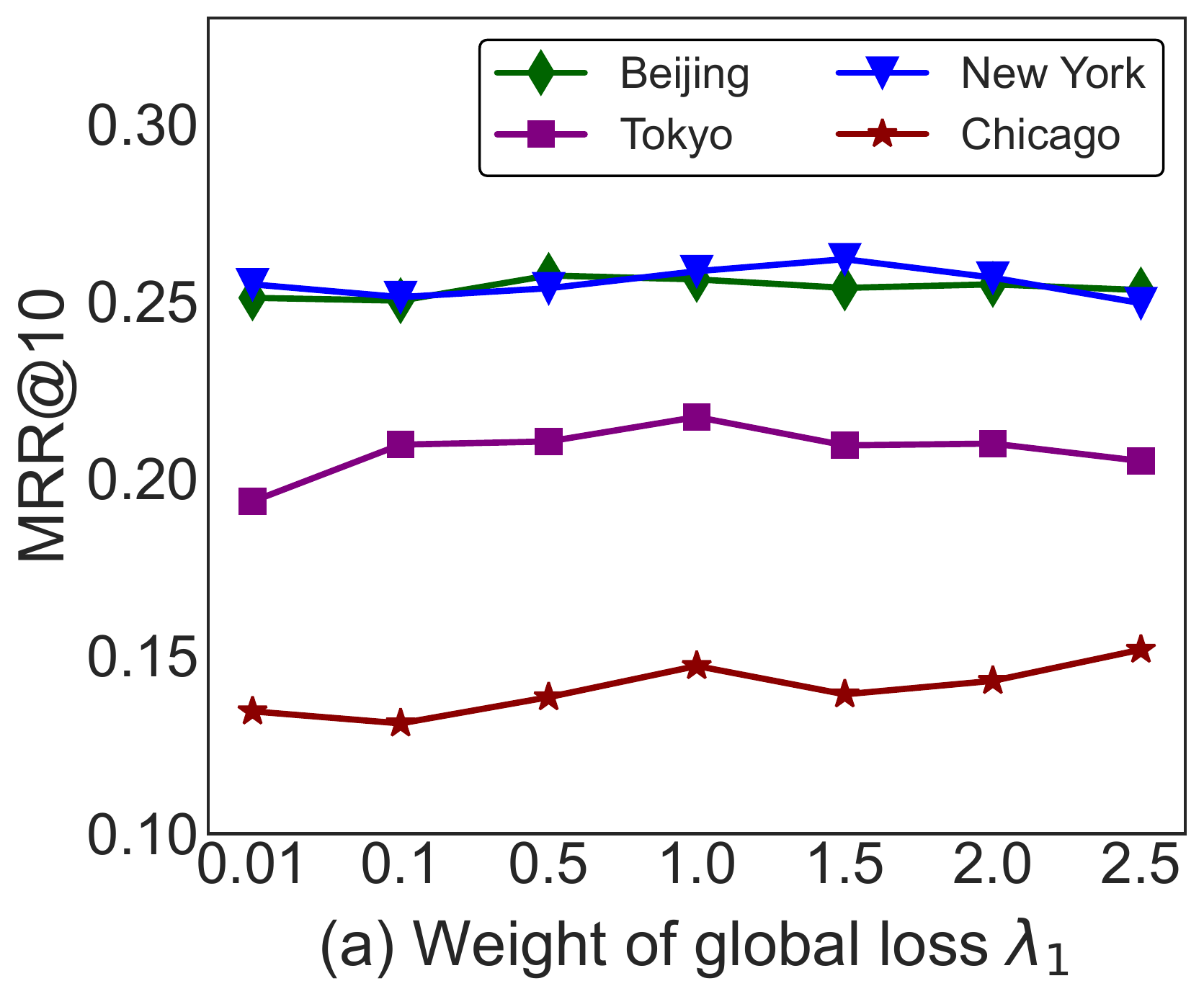}}
      \subfigure{
    \label{fig-para-local}
    \includegraphics[width=0.31\columnwidth]{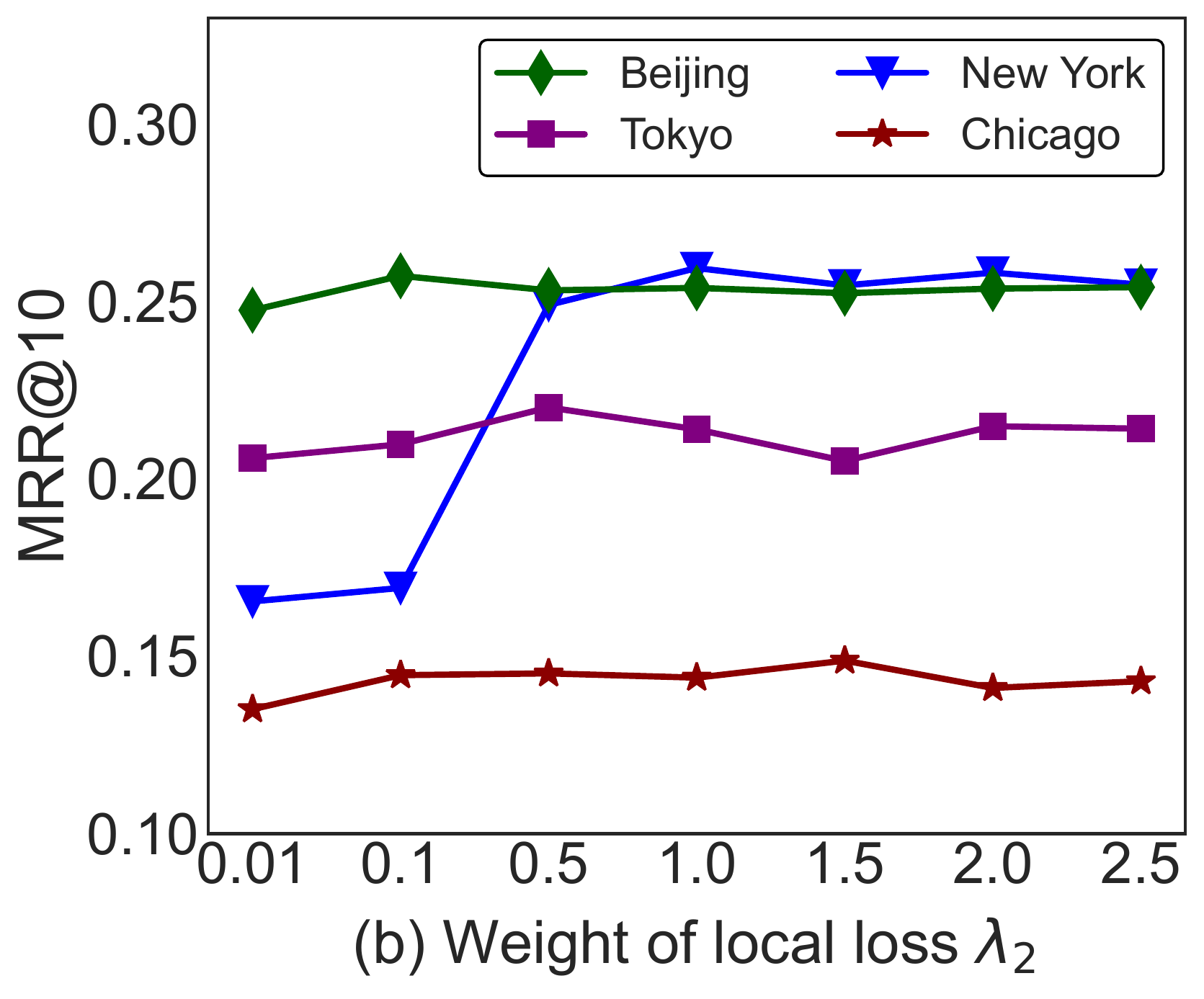}}
  \subfigure{
    \label{fig-para-neighbor} 
    \includegraphics[width=0.31\columnwidth]{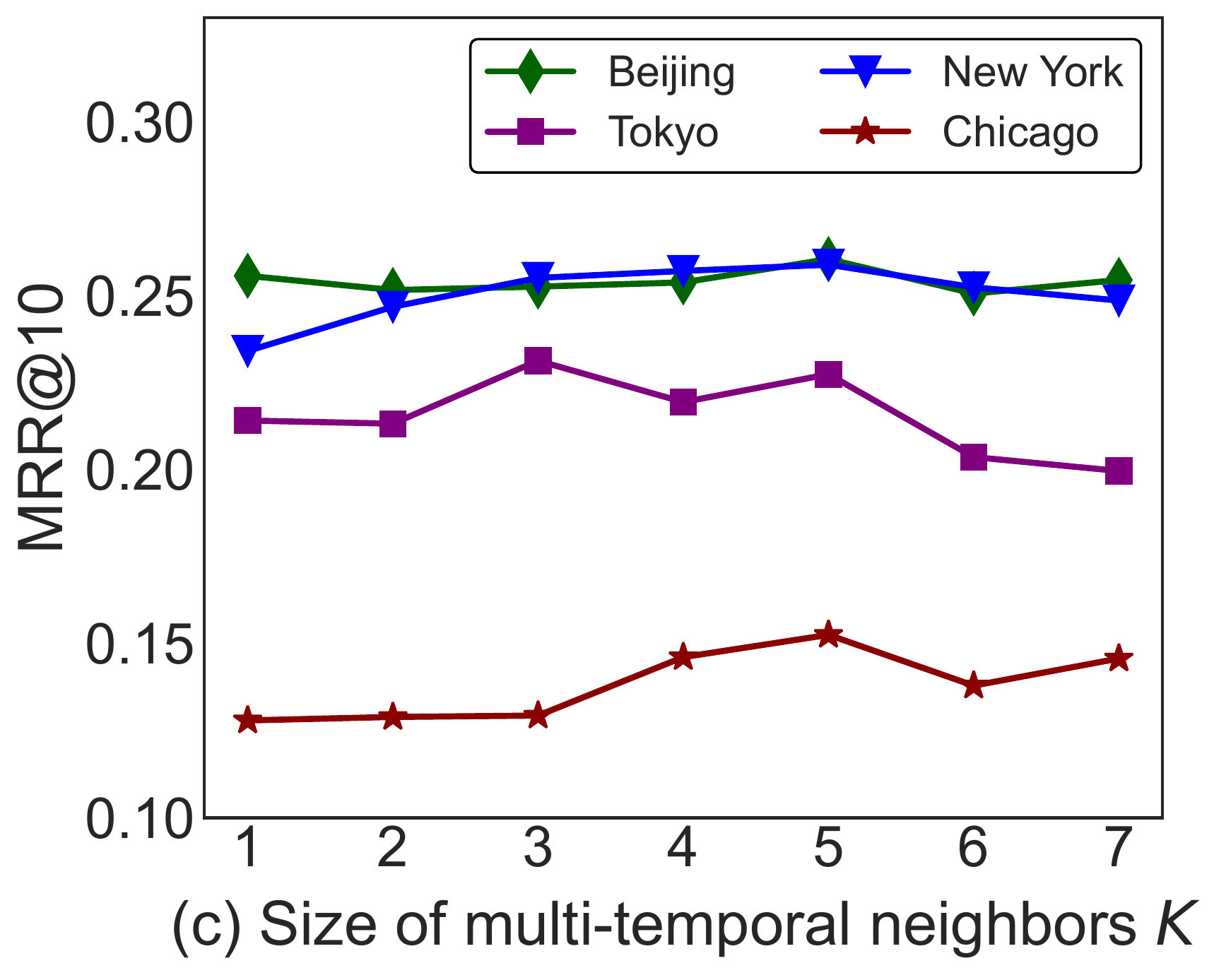}}
  \vspace{-2ex}
  \caption{Parameter analysis on four citywide datasets.}
  \vspace{-3ex}
  \label{fig-parameter}
\end{figure}

As depicted in Figure \ref{fig-para-global} and \ref{fig-para-local}, we first study the weight of global loss $\lambda_1$ and local loss $\lambda_2$ respectively. When increasing the coefficient for each loss\hide{ function}, the results slightly get better and keep stable in general, with the exception that the performance on New York increases rapidly \hide{at the smaller weights}in the beginning and then remains at high scores. This indicates integrating more local evolving relationship information is necessary for some scenarios. Overall, our collaborative global-local learning makes the model training more expressive and stable.
\vspace{-4ex}
\licom{
\subsubsection{Scales of spatially evolving context} We also investigate the number of sampled neighbors $K$ for capturing critical multi-temporal spatial factors in \tconv. Figure \ref{fig-para-neighbor} shows that the performance of \model gradually improves at first and then declines with \hide{involved}\hide{sampled} neighbors growing, which achieves the best at $K=5$. The reason is that communicating more location neighbors in \tconv can provide more informative spatial contexts, while sampling excessive neighbors may lead to unexpected homogenization of the propagation. To conclude, moderate fine-grained contextual information can help to outperform all baselines in our problem \problem.
}
\vspace{-0ex}
\section{Conclusion}
In this paper, we provided a new multi-temporal perspective to understand the location relationship, which is significant in urban intelligence. In detail, we proposed a spatially evolving GNN framework, named \model, to effectively discover multi-temporal location relationships in urban areas. The designed spatially evolving convolution can capture intra- and inter-time spatial contexts with dynamic influence. To overcome the issue of data sparsity, we also devised essential self-supervised learning tasks to integrate evolving patterns. Extensive experiments were conducted on four datasets to \hide{comprehensively }verify the effectiveness of the \liu{proposed} model.

\begin{acks}
The work was partially supported by grants from the National Natural Science Foundation of China (Grant No.61960206008).
\end{acks}

\normalem
\bibliographystyle{ACM-Reference-Format}
\balance
\bibliography{ref}

\newpage
\nobalance
\appendix
\section{Appendix}
\label{appendix}
\hide{In the appendix, we first \hide{comprehensively }introduce the relationship datasets. Then more experimental details and results are given for reproducibility.
\vspace{-0.5ex}}
\subsection{Dataset Introduction}
\label{A-dataset}
\licom{
\begin{itemize}[leftmargin=*,topsep=3pt]
\item \B{Business-based Relationships.} We first use the QueryBJ dataset collected from Baidu Maps, which contains millions of map search query logs \revise{from January 2019 to August 2019} in \textbf{Beijing}. Each query log provides the user's query session with timestamps. Following the works \cite{chen2021points,liu2020decoupled}, the two types of relations between locations $v_i$ and $v_j$ are defined as: 
    
    \underline{(1)} Users viewed $v_i$ also viewed $v_j$ within a query session\hide{ (\textit{competitive}}; 
    
    \underline{(2)} Users viewed $v_i$ then viewed $v_j$ across different \hide{query }sessions\hide{(\textit{complementary})}.
    
    As for the public Foursquare dataset \cite{yang2014modeling}, we collect the check-in data \hide{from the representative city}\revise{from April 2012 to February 2013 in} \textbf{Tokyo}. All visited locations of a user are classified into several categorical trips (e.g., all restaurants visited in a day form a categorical trip). Similarly, \hide{inspired by the work}following \cite{li2020competitive}, we have the two kinds of relationships as follows:
    
    \underline{(3)} Users visited $v_i$ also visited $v_j$ within a categorical trip\hide{ (\textit{competitive})};
    
    \underline{(4)} Users visited $v_i$ then visited $v_j$ across different categories\hide{ categorical trips}\hide{(\textit{complementary}}.  

    According to these \pending{relationship} studies \cite{mcauley2015inferring, li2020competitive, chen2021points}, we refer to (1) and (3) as \textit{competitive} relationships while the other (2) and (4) are \hide{considered}known as \textit{complementary} relationships.
    \item \B{Mobility-based Relationships.} For the taxi driving data, the public NYCTaxi\hide{ \footnote{https://nyc.gov/site/tlc/about/tlc-trip-record-data.page}} includes the order records traveling throughout \textbf{New York}\hide{ City} \revise{from January 2015 to June 2015}, while the \hide{opening }dataset DivvyBike\hide{\footnote{https://ride.divvybikes.com/system-data}} collects bike riding orders from \revise{January 2017 to June 2017}\hide{the bike sharing system Divvy}\hide{of people daily using} in \textbf{Chicago}. In this scene, the location can be a bike station or a region. Each trajectory by taxi or bike includes the pick-up and drop-off locations with timestamps, which connects a pair of locations\hide{ $(v_i,v_j)$}. After counting the overall records for mobile-based relevancy\hide{ $fr_{ij}$}, we label the two meaningful relationships between location pairs, namely \textit{high-flow} (top 25\% high) and \textit{low-flow} (top 50\% high) relationships according to the mobility degree.
\end{itemize}
\hide{Note that the behavior and trajectory information is often insufficient and can not cover the majority of locations in practical scenarios. Thus, the observed relationships are quite sparse, meaning that there are more valuable hidden relationships to be discovered by machine learning methods.}
}
\vspace{-0.5ex}
\subsection{Experiment Details}
\label{A-detail}
\subsubsection{Implementation and setup} We train the model on 24 Intel CPUs and a group of Tesla P40 GPUs. \pending{The unified input location features are initialized with one-hot embedding for generalized representation learning.} \hide{The code is available at xxx.}\licom{For data splitting, we guarantee that the testing and validation sets $\mathcal{Y}_{pred}$ are absolutely independent of the training set $\mathcal{Y}_{trn}$ across all times, meaning that every relational edge in $\mathcal{Y}_{pred}$ at a certain time will not appear in $\mathcal{Y}_{trn}$ at any time \hide{segment}.}
\subsubsection{Evaluation Metrics}
\label{A-metric}

\revise{In inferring location relationships, the model should be able to accurately rank the relational locations instead of only predicting whether the relationships exist or not. Thus, it is important that relevant locations are ranked higher than irrelevant ones with the ranking metrics MRR and HR. Given the test set $T$, we follow these steps for evaluation at each time segment: (1) For each relational pair $(v_i,v_j,r)$, we first calculate all pairwise scores for relation $r$ between $v_i$ and the $N$ locations in the candidate set $S$ using the relational-specific prediction function $f_r$. (2) We sort these $N$ locations in descending order and obtain the ranked list. The ranking index of node $v_j$ is denoted as $rank_j$ (i.e., a certain position in the list). (3) If $rank_j \le k$, we consider it a hit (successfully discovering the target location $v_j$ from the top-K locations), otherwise, we consider it a miss. The HR@k metric is defined as the average of total hits over the entire test set $T$, which is denoted as $HR@k=\sum_{rank_j\le k} \frac{1}{|T|}$. (4) To calculate the MRR@k metric, we first compute the reciprocal rank, denoted as $\frac{1}{rank_j}$. Note that the reciprocal rank is 0 when getting a miss (i.e., $rank_j > k$). Thus, the MRR@k metric is defined as the average of total reciprocal ranks over the entire test set $T$, which is denoted as $MRR@k=\frac{1}{|T|}\sum_{rank_j\le k}\frac{1}{rank_j}$.
}

\hide{
We first detail the used metrics in our experiment \licom{to comprehensively evaluate the performance of our model and baselines in discovering multi-temporal location relationships}.
\textit{MRR@k} and \textit{HR@k} are mathematically defined as:
\begin{equation}
\small
\label{eq-metric}
\begin{split}
    \pending{MRR@k} &= \frac{1}{M}\sum_{i=1}^M \mathcal{I}_{rank_i\le k}\cdot\frac{1}{rank_i},\\
   \hide{MRR@k &= \frac{1}{k}\sum_{i=1}^k \mathcal{I}_{rank_i\le k}\cdot\frac{1}{i}\\}
    HR@k &= \frac{1}{M}\sum_{i=1}^M \mathcal{I}_{rank_i\le k},
\end{split}
\end{equation}
where $M$ is the size of testing set, $rank_i$ represents the ranking index of the location $v_i$, $\mathcal{I}$ is the indicator function ($\mathcal{I}_{rank_i\le k}=1$ iff $v_i$ is among the top $k$ and 0 others).
}

\subsubsection{Parameter Settings}
For all models, we set the dimension of embedding and hidden layers to 64 with the two-layer GNN architecture. Since some baselines are not proposed for relationship discovery, the DistMult function \cite{yang2015embedding} is also adopted as the final prediction layer for these models. In our proposed \model, the model is trained by Adam optimizer with an initial learning rate of 0.01. The number of negative samplers is set to 5. For the GNN framework of \gnn, we set the number of distance bins to 40, and the size of multi-temporal neighbors to 5 for spatial context learning. For the self-supervised learning of \ssl, the related parameters of balancing weights (i.e., $\lambda_1$ and $\lambda_2$) and grdding size in the global loss are set according to the experimental results on the validation set (The \hide{searching space}certain range is shown in Figure \ref{fig-parameter} and Figure \ref{A-parameter}).

For baseline models,\hide{ we tune the parameters based on default settings in the paper to get optimal performance for each model. Specifically,} the number of random walk and the path length are both set to 5 in pathGCN, while the composition operator of subtraction is used with the dropout rate of 0.2 in CompGCN. For high-order graph learning methods, the set of integer adjacency powers in MixHop is defined as $P=\{0,1,2\}$, and the kernel size in NL-GNN is set to 5 for effective non-local context learning. As to graph self-supervised learning methods, we also use the bilinear scoring function as the discriminator for DGI. The degree centrality function with the dropping probabilities of 0.3 and 0.2 is employed for graph augmentation in GCA. The balancing coefficient for relational self-supervised learning is set to 0.9 in RGRL. For dynamic GNN models, the MLP and GRU modules are adopted for recurrent updater in EvolveGCN and ROLAND. For relationship learning methods (DecGCN and IRGNN), we stack two convolutional layers with two 3-layer MLP functions for complex relational dependencies. For DeepR, The grid size of buckets and the number of sectors are set to 100 (meters) and 4, respectively. The number of attentive heads is set to 2 with the scaling factor of 2 in PRIM.

\subsubsection{Baseline Method Description}
\label{A-baseline}
We compare \model \hide{model }with the following methods for multi-temporal relationship inference\hide{ among locations in the urban area}:
\label{a-baseline}
\begin{itemize}[leftmargin=*,topsep=3pt]
    \item \B{GCN} \cite{kipf2017semi} is a well-known graph neural network, which aggregates nodes with topological weights for relational modeling.
    \item \B{PathGCN} \cite{eliasof2022pathgcn} adopts the point-wise graph convolutions to learn the complex spatial operator from random paths for improving relationship prediction performance. 
    \item \B{CompGCN} \cite{vashishth2020composition} extends the GCN architecture to jointly embeds both nodes and relationships in the \hide{relational }graph. It can incorporate multi-relational information with composition operations.
    \item \B{MixHop} \cite{abu2019mixhop} is a kind of higher-order message passing \hide{GNN model}network, where nodes receive abundant and distant information with mixing feature representations of neighbors at various distances.
    \item \B{NL-GNN} \cite{liu2021non} is a recent non-local aggregation framework with an attentive sorting to capture global relational structures.
    \item \B{DGI} \cite{velickovic2019deep} leverages the  mutual information maximization for graph self-supervised learning (SSL) through a local-global scheme.
     \item \B{GCA} \cite{zhu2021graph} develops the graph contrastive SSL with advanced adaptive augmentations to enhance representation learning.
     \item \B{RGRL} \cite{lee2022relational} is a recent relational-aware SSL framework to alleviate the data scarcity issue by considering the relationship among nodes in both global and local perspectives.
     \item \B{EvolveGCN} \cite{pareja2020evolvegcn} employs an RNN module to dynamically update weights of internal GNNs\hide{, which allows the GNN model to make} for dynamic link predictions.
     \item \B{ROLAND} \cite{pareja2020evolvegcn} is the latest snapshot-based GNN model to further generalize the relational GNN to a dynamic setting, which can enable the prediction of multi-time relations.
     \item \B{DecGCN} \cite{liu2020decoupled} generates node embeddings in separated relationship-specific spaces, which can capture the mutual inference between structural and semantic information for relationship discovery.
     \item \B{IRGNN} \cite{liu2021item} is proposed to discover multiple relationships by incorporating multi-hop relational information on sparse graphs.
     \item \B{DeepR} \cite{li2020competitive} introduces spatial adaptive GNN model to handle the unique spatial attribute of location graphs and achieves great performance for static relationship inference.
     \item \B{PRIM} \cite{chen2021points} is the current state-of-the-art GNN model for location relationship inference. Boht the weighted relational convolution \hide{module }and self-attentive spatial context extraction improve the \hide{prediction }results.
\end{itemize}
\vspace{-1.5ex}
\hide{
\begin{figure}
\setlength{\abovecaptionskip}{2.mm}
\setlength{\belowcaptionskip}{-0.cm}
  \centering
  \subfigure{
    \label{fig-ratio-bj} 
    \includegraphics[width=0.48\columnwidth]{figure/ratio/train_ratio_bj-new.pdf}}
      \subfigure{
    \label{fig-ratio-tky}
    \includegraphics[width=0.48\columnwidth]{figure/ratio/train_ratio_tky-new.pdf}}
 \\[-3ex]
  \subfigure{
    \label{fig-ratio-nyc} 
    \includegraphics[width=0.48\columnwidth]{figure/ratio/train_ratio_nyc-new.pdf}}
    \subfigure{
    \label{fig-ratio-chi} 
    \includegraphics[width=0.48\columnwidth]{figure/ratio/train_ratio_chi-new1.pdf}}
  \vspace{-3mm}
  \caption{Sparsity analysis on four citywide datasets.}
  \vspace{-4.5mm}
  \label{A-ratio}
\end{figure}
}

\begin{table}
	\centering
    \caption{Time-specific performance evaluation (MRR@10) over \hide{segments }times on two datasets of Business-RD and Mobi-RD.}
    \label{table-time-exp}
    \vspace{-4mm}
	\scalebox{0.85}{\begin{tabular}{c|c|cccc}
		\toprule
		  \B{Dataset} & \B{Method} & \B{Morning} & \B{Midday} & \B{Night} & \B{Midnight} \\
		\midrule
            \multirow{7}{*}{\shortstack{\textsc{Beijing}}} 
            & NL-GNN & 0.2302 & 0.1803 & 0.1821 & 0.1340 \\
            & RGRL & 0.2232 & 0.1901 & 0.2018 & 0.1352\\
            & EvolveGCN & \underline{0.2718} & 0.2072 & 0.2169 & \underline{0.1760} \\
            & ROLAND & 0.2402 & 0.2080 & \underline{0.2187} & 0.1567 \\
            & DeepR & 0.2152 & \underline{0.2247} & 0.2159 & 0.1334 \\
            & PRIM & 0.2114 & 0.2005 & 0.1962 & 0.1409 \\
            & \model & \B{0.3206} & \B{0.2442} & \B{0.2621} & \B{0.2602} \\
            \midrule
            \multirow{7}{*}{\shortstack{\textsc{New York}}} 
            & NL-GNN & 0.1386 & 0.2042 & 0.2029 & 0.1532 \\
            & RGRL & 0.1778 & 0.1419 & 0.1607 & 0.1850\\
            & EvolveGCN & 0.1463 & 0.2185 & 0.2259 & 0.1576 \\
            & ROLAND & 0.1799 & 0.2027 & 0.2129 & \underline{0.1993} \\
            & DeepR & 0.1838 & 0.2122 & 0.2080 & 0.1734 \\
            & PRIM & \underline{0.2072} & \underline{0.2276} & \underline{0.2420} & 0.1976 \\
            & \model & \B{0.2722} & \B{0.2367} & \B{0.2714} & \B{0.2415} \\
		\bottomrule
	\end{tabular}}
	\vspace{-5mm}
\end{table}

\subsection{Additional Experimental Results}
\label{A-exp}

\subsubsection{\B{Time-specific Performance Analysis}}

Table \ref{table-time-exp} exhibits the multi-temporal relationship prediction results over all times in a day. The proposed \model provides stable performance gains over \revise{major competitive} baselines at all time periods, which consistently verifies the superiority of our multi-slot spatial learning framework in time-aware predictions. We also observe that most baselines perform worse at midnight due to the sparsity of location relationships. We \liu{observe} that the graph at midnight with fewer people activities only contains limited environmental information for location relationship inference. Moreover, we \liu{find} that dynamic GNNs (EvolveGCN and ROLAND) can partly alleviate the above issue thanks to the capability of involving the relational knowledge from the previous time, which agrees with the fact that considering dynamic relationships is essential and valuable. However, the \hide{time-specific }performance is still unsatisfactory. By contrast, our model can achieve stable and accurate prediction results at any time, demonstrating the exhaustive effectiveness of \model.

\hide{
\subsubsection{\B{Relationship Sparsity Analysis}\hide{Robustness Analysis of Relationship Sparsity}}
\label{A-sparsity-exp}
\hide{As discussed in Section \ref{sec-ssl}, the proposed spatially evolving self-supervised learning (\ssl) scheme is capable of handling the issue of relationship sparsity.}\hide{ While \ssl has been proven to be effective in Section \ref{sec-abla},} Since the sparsity of location relationships is a frequent issue in practice and has an impact on the performance of relationship discovery methods, we further evaluate the effectiveness of \model and \hide{several}major competitive baselines as the degree of relational sparsity increases. As presented in Figure \ref{A-ratio}, when decreasing the ratio of training relationships (i.e., removing more percentage of relational edges in the graph), the performance gradually declines due to fewer\hide{sparser} labeled relationships. \hide{The comparison between \model and other powerful baselines shows}The results show that the proposed \model can consistently outperform other powerful methods under all sparse settings. More importantly, our model sometimes further expands the prediction advantage when facing much sparser data\hide{under the much sparser settings}, which demonstrates that \model is more robust to the relationship sparsity problem.
\pending{Overall, the experiments additionally validate the robustness and effectiveness of spatially evolving learning framework to alleviate the data sparsity problem for location relationship discovery.}
}

\begin{table}
    \centering
    \caption{The efficiency studies on Beijing dataset.}
    \label{table-efficiency-exp}
    \vspace{-4mm}
	\scalebox{0.9}{\begin{tabular}{ccc}
		\toprule
		  \B{Method} & \B{Training (s/epoch)} & \B{Inference (s/rank)} \\
		\midrule
            RGRL & 1.4464 & 0.1212 \\
            ROLAND & 2.8150 & 0.1540 \\
            DeepR & 2.3462 & 0.2340 \\
            PRIM & 1.8416 & 0.2058 \\
            \model & 2.1631 & 0.2517 \\
		\bottomrule
	\end{tabular}}
	\vspace{-5mm}
\end{table}

\begin{figure}
\setlength{\abovecaptionskip}{2.mm}
\setlength{\belowcaptionskip}{-0.cm}
  \centering
  \subfigure{
    \label{fig-para-gridlen} 
    \includegraphics[width=0.75\columnwidth]{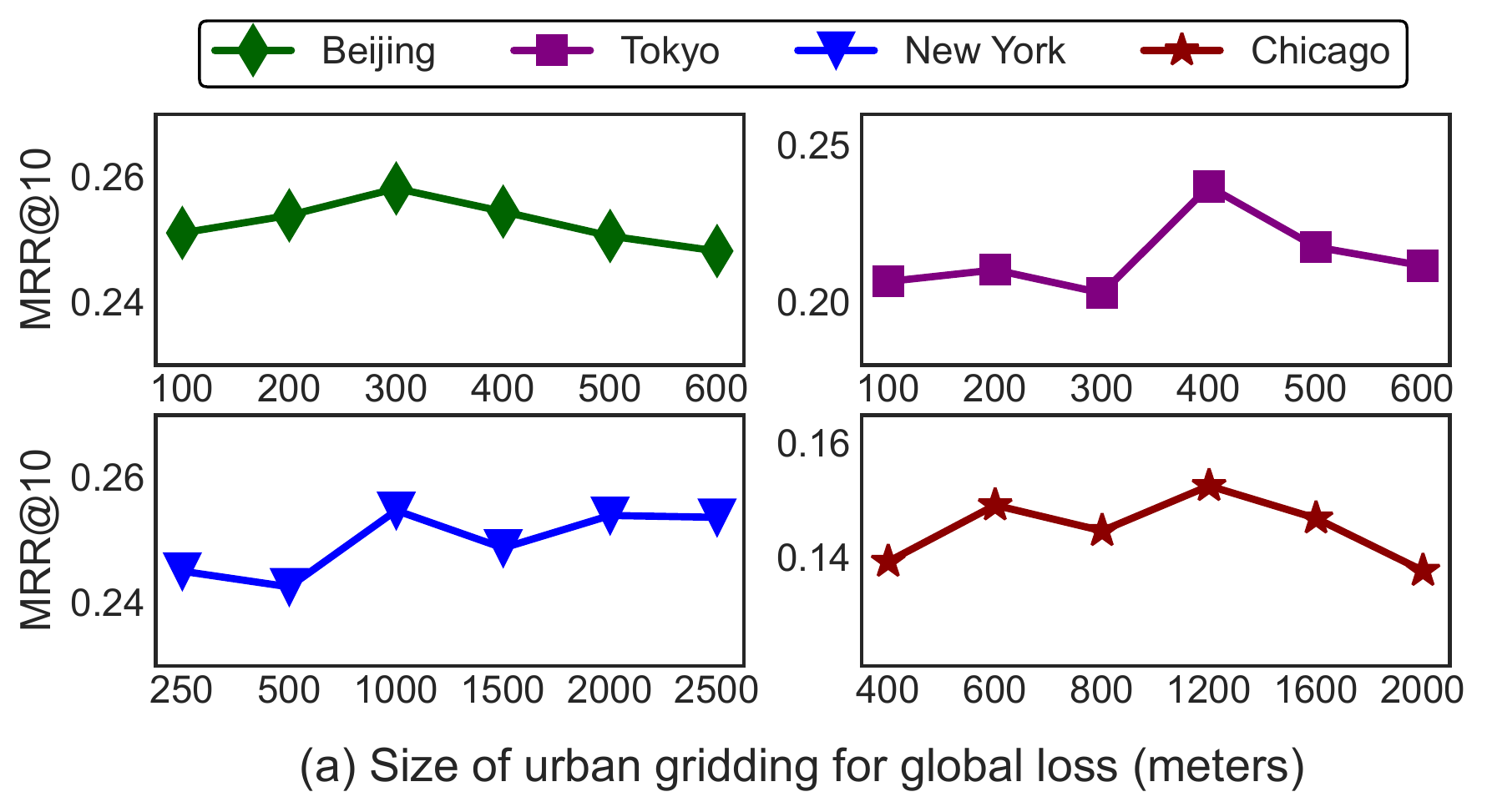}}
     \\[-3ex]
      \subfigure{
    \label{fig-para-sampling}
    \includegraphics[width=0.4\columnwidth]{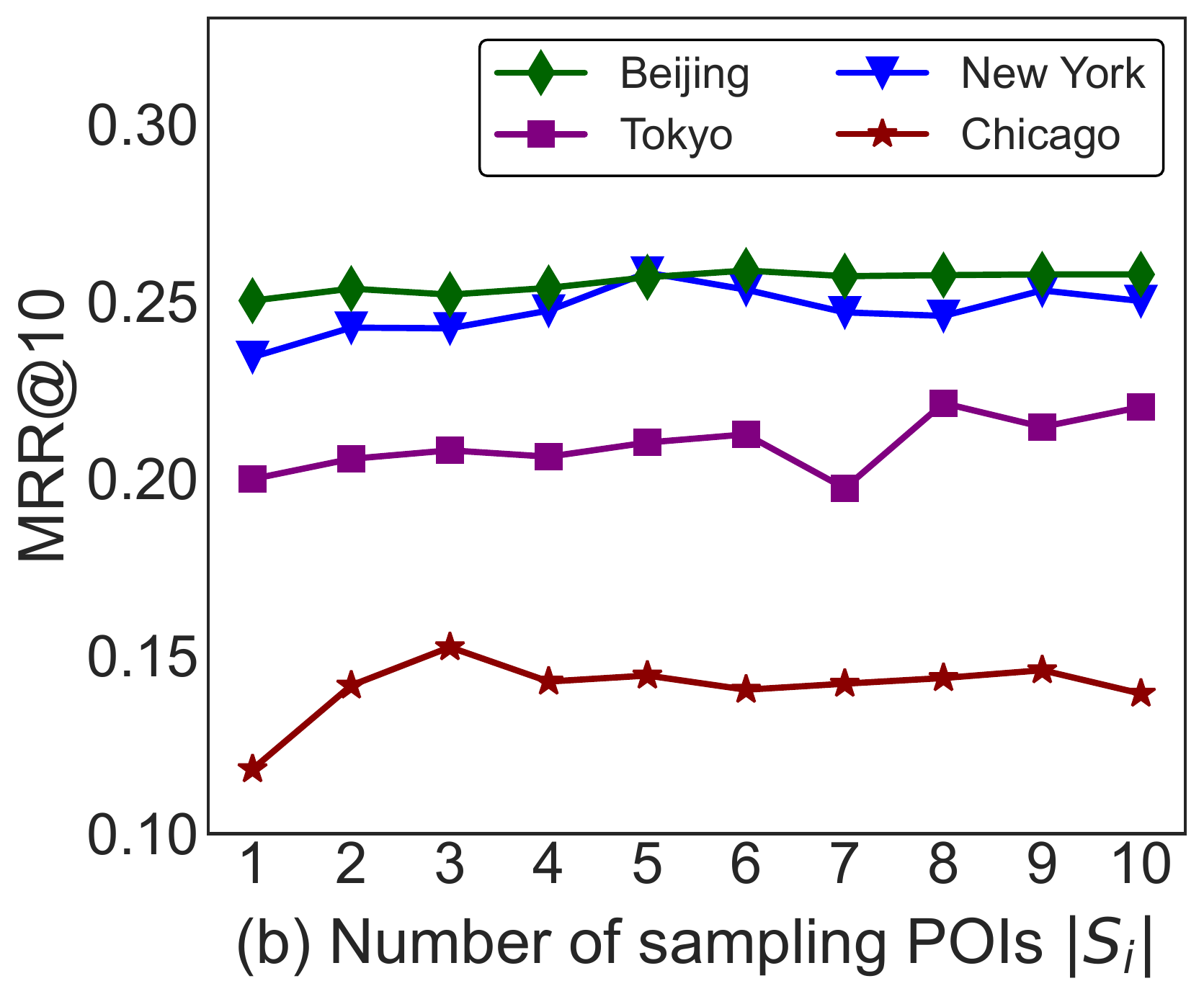}}
    \subfigure{
    \label{fig-para-bin}
    \includegraphics[width=0.4\columnwidth]{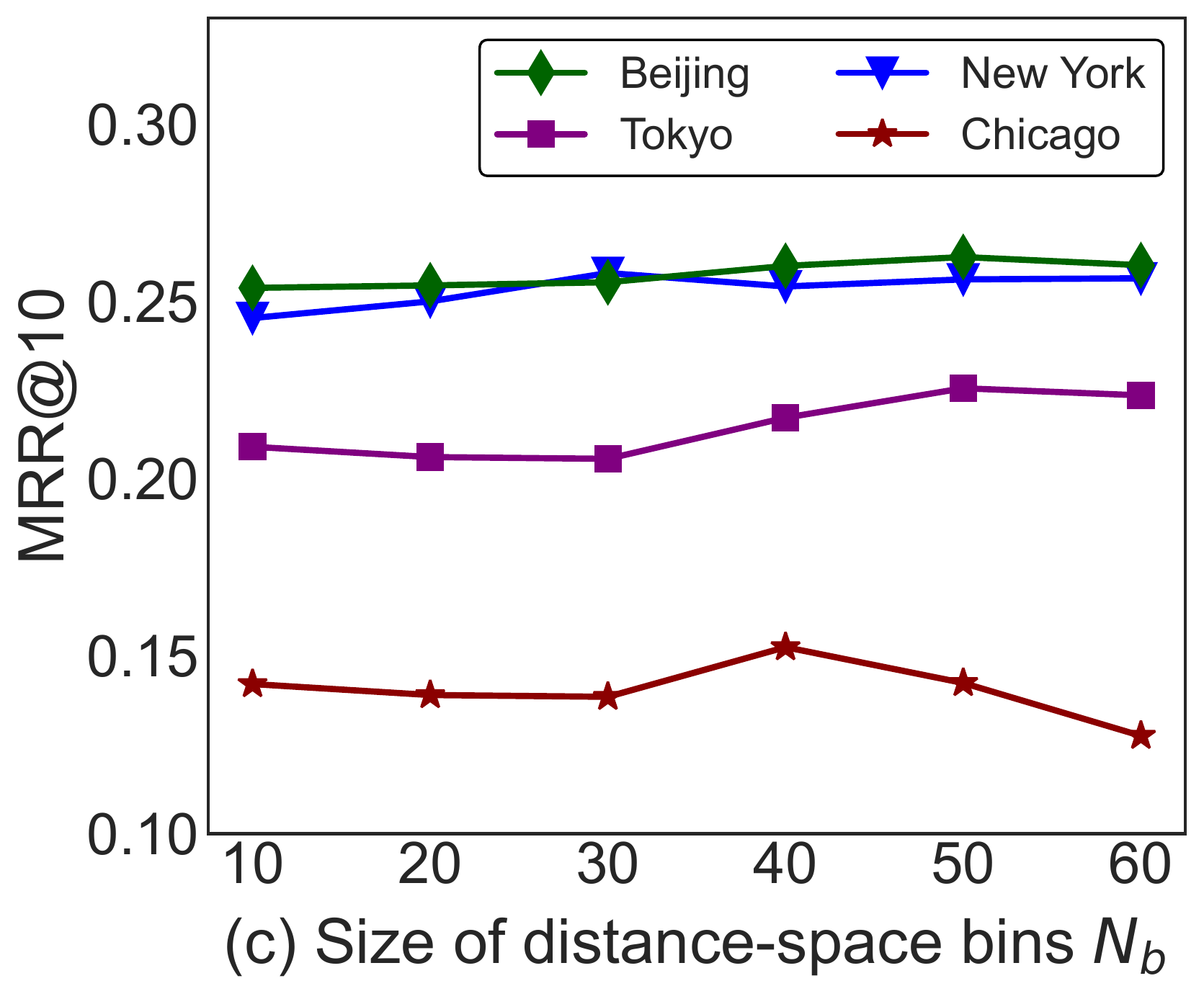}}
  \vspace{-3mm}
  \caption{More parameter analysis on four citywide datasets.}
  \vspace{-4mm}
  \label{A-parameter}
\end{figure}

\revise{
\subsubsection{\B{Efficiency Analysis}}
We also conduct the efficiency evaluation for the training and inference time. We compare our model with several most competitive baselines on the largest Beijing dataset in Table \ref{table-efficiency-exp}. The results show that \hide{our model }\model can be comparably efficient with other models and will not sacrifice much computation and training time to trade for performance. It is worth noting that the “inference (s/rank)” \hide{in the table }refers to the total time of ranking all nodes on the graph for a specific location. When we use \model to discover location relationships, inferring the ranking list for one location on a large graph (over 30,000 nodes) takes only 0.25s on average. 
}

\subsubsection{\B{Additional Parameters Analysis}}
\label{A-para-exp}
\hide{Size of urban gridding. }As shown in Figure \ref{A-parameter}, we first present the influence of urban gridding size in the global spatial information maximum loss, which determines the spatial granularity for global region pooling. With the growth of the gridding size (e.g., the larger region area), the performance of our model first increases and then tends to decrease. The reason is that \model needs a suitable splitting scale for higher-level aggregation according to different citywide application domains. While the larger region grid can contain more informative spatial locations, additional redundancies may be introduced for relationship learning. Moreover, we change the another important parameter of \ssl, i.e., geographical sampling size in $S_i$, to investigate the affect of heuristic negative sampler \licom{in Figure \ref{fig-para-sampling}}. The performance improves slightly at first and then keep relatively stable as the number of negative samples grows, verifying that more grid-based negative samples can enhance the global spatial self-supervised learning. \licom{Finally, Figure \ref{fig-para-bin} presents the effect of the size of distance-space bins $N_b$. We can observe that our model can stably perform well with changing distance bins due to the adaptive ability to learn spatial distances in \model.}

\end{document}